\documentclass[10pt,twocolumn,letterpaper]{article}

\usepackage{cvpr}              

\usepackage{graphicx}
\usepackage{amsmath}
\usepackage{amssymb}
\usepackage{booktabs}

\usepackage[pagebackref,breaklinks,colorlinks]{hyperref}

\usepackage{epsfig}
\usepackage{multirow}
\usepackage{multicol}
\usepackage{wrapfig,lipsum,booktabs}
\usepackage{caption}
\usepackage{environ}
\usepackage[table,dvipsnames]{xcolor}
\definecolor{aliceblue}{rgb}{0.94, 0.97, 1.0}

\newcommand{\CC}[1]{\cellcolor{blue!#1}}

\usepackage[capitalize]{cleveref}
\crefname{section}{Sec.}{Secs.}
\Crefname{section}{Section}{Sections}
\Crefname{table}{Table}{Tables}
\crefname{table}{Tab.}{Tabs.}

\def\cvprPaperID{6431} 
\def\confName{CVPR}
\def\confYear{2022}

\begin{document}

\twocolumn[{
\begin{@twocolumnfalse}

\title{Weakly-supervised Action Transition Learning for \\Stochastic Human Motion Prediction}

\author{
Wei Mao$^1$, \;\;Miaomiao Liu$^{1}$,\;\; Mathieu Salzmann$^{2,3}$\;\; \\
$^1$Australian National University; $^2$CVLab, EPFL; $^3$ClearSpace, Switzerland\\
{\tt\small \{wei.mao, miaomiao.liu\}@anu.edu.au,}\;\;{\tt\small mathieu.salzmann@epfl.ch}
}
\maketitle

\begin{center}
\setlength\tabcolsep{1pt}
\vspace{-0.8cm}\begin{tabular}{c}
      \includegraphics[width=0.8\linewidth]{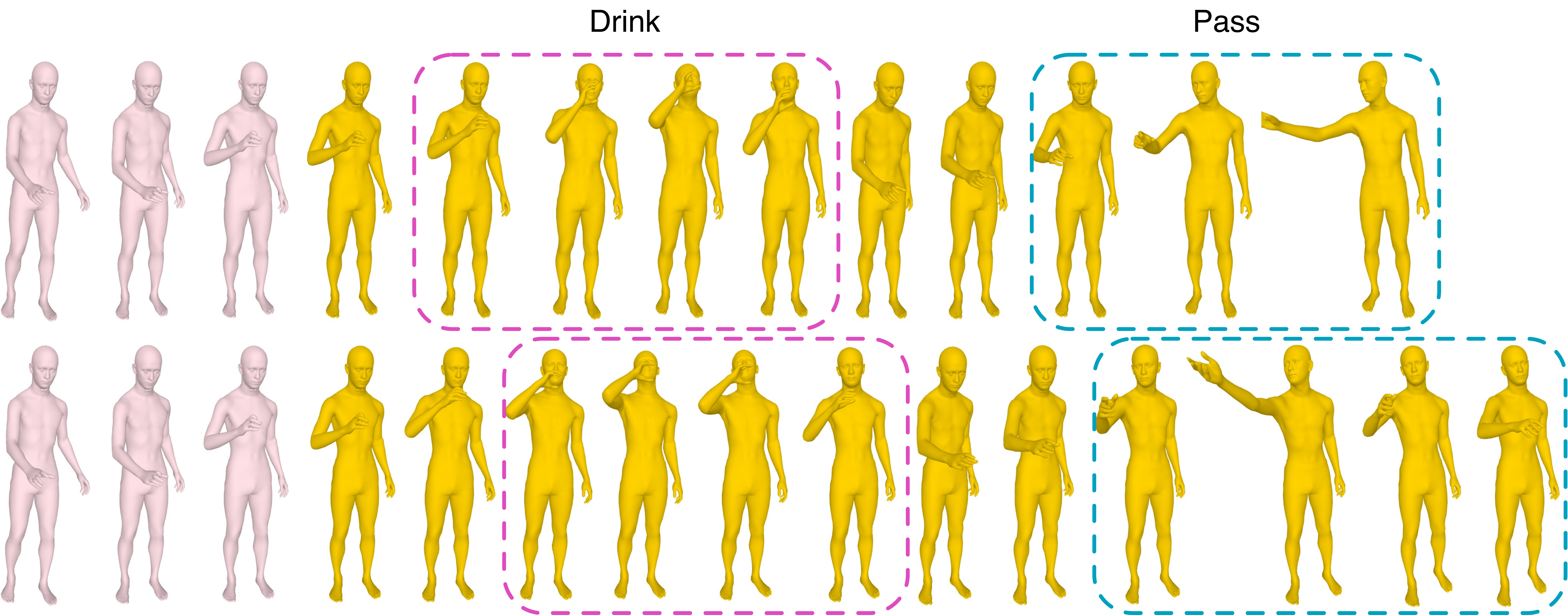}
\end{tabular}
\end{center}
\vspace{-0.8cm}
\captionof{figure}{{\bf Action-driven stochastic human motion prediction.} Given a past motion (pink) and a sequence of future action labels, our model generates action-specific future poses (yellow). We show two different futures generated for the same history and actions. Our model allows these predictions to have different lengths. The motions are down-sampled to the same frame rate for visualisation.
\vspace{0.2cm} }\label{fig:opening}
\end{@twocolumnfalse}
}]

\begin{abstract}
\vspace{-0.5cm}
  We introduce the task of \emph{action-driven stochastic human motion prediction}, which aims to predict multiple plausible future motions given a sequence of action labels and a short motion history. This differs from existing works, which predict motions that either do not respect any specific action category, or follow a single action label. In particular, addressing this task requires tackling two challenges: The transitions between the different actions must be smooth; the length of the predicted motion depends on the action sequence and varies significantly across samples. As we cannot realistically expect training data to cover sufficiently diverse action transitions and motion lengths, we propose an effective training strategy consisting of combining multiple motions from different actions and introducing a weak form of supervision to encourage smooth transitions. We then design a VAE-based model conditioned on both the observed motion and the action label sequence, allowing us to generate multiple plausible future motions of varying length. We illustrate the generality of our approach by exploring its use with two different temporal encoding models, namely RNNs and Transformers. Our approach outperforms baseline models constructed by adapting state-of-the-art single action-conditioned motion generation methods and stochastic human motion prediction approaches to our new task of action-driven stochastic motion prediction. Our code is available at \url{https://github.com/wei-mao-2019/WAT}.
\end{abstract}
\vspace{-0.5cm}
\section{Introduction}\label{sec:intro}
Modeling human motion has broad applications in human-robot interaction~\cite{koppula2013anticipating}, virtual/augmented reality~(AR/VR)~\cite{starke2019neural} and animation~\cite{van2010real}. 
As such, it has been an active research problem for many years~\cite{brand2000style}. In
particular, recently, great progress has been made in predicting future motion given an observed past motion sequence~\cite{aliakbarian2020stochastic,yuan2020dlow}. Addressing this could have a significant impact on autonomous systems, allowing them to forecast potential dangers and plan their actions accordingly.
Nevertheless, except for a few early methods that predict motions of a single action category~\cite{fragkiadaki2015recurrent,JainZSS16},
recent methods~\cite{mao2020history,yuan2020dlow} mostly focus on action-agnostic predictions.  
Thus, they cannot be used by an autonomous system to generate specific potential future scenarios encoded by a sequence of action labels, for example to evaluate the consequences of a person on a sidewalk either \emph{walking} to the crossing, \emph{waiting} for the green light, and \emph{crossing} the road, or instead \emph{running} on the street and \emph{stopping} in front of the car.
By contrast, recent works on human motion synthesis can generate action-specific sequences~\cite{guo2020action2motion,petrovich21actor}. However, these methods neither leverage past motion observations, nor synthesize transitions between different actions.
In this work, we therefore introduce the task of \emph{action-driven stochastic human motion prediction}, which aims to predict a set of future motions given a sequence of action labels and past motion observations.

One of the key challenges of this task arises from the fact that humans can perform
motions with all kinds of action transitions. For example, when one \emph{walks} to
a table, they can then either \emph{grab} a drink, or \emph{sit} on a chair, or
\emph{place} something on the table, or perform any combination of the above.
Constructing a dataset that covers this huge space of possible action transitions  is therefore virtually impossible, significantly complicating training a model for this task. As a matter of fact, to the best of our knowledge, almost all human motion datasets contain sequences that depict a single action. While the recent BABEL dataset~\cite{BABEL:CVPR:2021} constitutes the only exception with multiple actions per sequence, it contains only a small subset of action transitions, which, as evidenced by our experiments, does not suffice to learn to generalize to arbitrary ones.

To tackle the diversity of human action transitions with such limited data, we develop a weakly-supervised training strategy that only relies on a motion smoothness prior. Specifically, we generate multi-action sequences by combining historical and future motions from different action categories, and account for the lack of supervision during the transition between two actions by simply encouraging the predicted motion to be temporally smooth. As will be shown by our experiments, such a simple prior suffices to model natural action transitions.

The second main challenge of our task arises from the stochastic nature of human motion: Several ways to perform one action sequence are equally plausible. To handle this stochasticity, we design a model based on a variational autoencoder~(VAE)~\cite{kingma2013auto}, conditioning the VAE on the observed past motion and on the action label sequence. We demonstrate the generality of this model by exploiting it with two different temporal encoding architectures, an RNN-based one
and a Transformer-based one.
Furthermore, to reflect the fact that some action sequences require more time to be executed than others, we introduce a simple yet effective strategy based on the prediction variance to produce multi-action motions of different lengths. This contrasts with most of the motion prediction literature, which predicts fixed-length motions, and, as illustrated by Fig.~\ref{fig:opening}, allows us to generate realistic, diverse future motions depicting the given actions in order and with varying length. We believe that our approach can also be beneficial for other tasks, e.g., music generation of variable length.

Our contribution can therefore be summarized as follows: 
(i) We introduce a new task, \emph{action-driven stochastic human motion prediction}, which bridges the gap between motion synthesis and stochastic human motion prediction; 
(ii) We propose a weakly-supervised training strategy to learn the action transitions without requiring an unrealistic amount of annotated data; 
(iii) We develop a simple yet effective way of predicting motions of varying length.

Our experiments on 3 human motion modeling benchmarks demonstrate the effectiveness of our approach, outperforming baseline models constructed by extending state-of-the-art action-conditioned motion synthesis methods and stochastic human motion prediction ones to our new task.
\vspace{-0.2cm}
\section{Related Work}
\noindent\textbf{Human Motion Prediction.} Most human motion prediction works~\cite{fragkiadaki2015recurrent,JainZSS16,Butepage_2017_CVPR,Martinez_2017_CVPR,pavllo_quaternet_2018,gui2018adversarial,LiZLL18,aksan2019structured,mao2019learning,wang2019imitation,gopalakrishnan2019neural,cai2020learning,mao2020history} 
focus on predicting human movements in a very short future~($<0.5$s). These
methods mainly differ in their temporal encoding strategies, using either recurrent
architectures~\cite{fragkiadaki2015recurrent,JainZSS16,Martinez_2017_CVPR,pavllo_quaternet_2018,gui2018adversarial,wang2019imitation,gopalakrishnan2019neural} or feed-forward
models~\cite{Butepage_2017_CVPR,LiZLL18,mao2019learning,aksan2019structured,cai2020learning,mao2020history}. Most
of them, however, do not aim to produce motions with the same past sequence that respect any given action label. The only
exceptions that incorporate action information are the early works
of~\cite{fragkiadaki2015recurrent,JainZSS16,Martinez_2017_CVPR,gui2018adversarial,gopalakrishnan2019neural}. However, such information is only used to help predict
future motion of the \emph{same} action as the historical motion. By contrast, we seek to predict
future motions of \emph{different and multiple} action categories.

Because, given one sequence of action labels, we aim to predict multiple plausible motions, which is more
closely related to diverse human motion
prediction~\cite{walker2017pose,lin2018human,barsoum2018hp,hernandez2019human,kundu2019bihmp,yan2018mt,aliakbarian2020stochastic,yuan2020dlow}.
To capture the distribution of future motions, these methods usually rely on deep generative models,
such as VAEs~\cite{kingma2013auto} and generative adversarial
networks~(GANs)~\cite{goodfellow2014generative}. Among the most recent ones, the work
of~\cite{aliakbarian2020stochastic} prevents the VAE from ignoring the random variables by perturbing
them; DLow~\cite{yuan2020dlow} focuses on learning the sampling process for diverse future predictions
from a pre-trained generative model. While these models indeed produce diverse and plausible future
motions, these generated motions do not follow any clear semantic categories. As such, they could not be
leveraged to help an autonomous system evaluate different scenarios defined in terms of semantic
human behaviors.
By contrast, our goal is to control the predicted future motion type using semantic information, i.e., a sequence of action labels. We therefore design a VAE-based model and a weakly-supervised training strategy that let us produce \emph{different} plausible future motions for the \emph{same} past motion and a sequence of action labels. 

\noindent\textbf{Human Motion Synthesis.} In contrast to motion prediction, human motion synthesis aims to generate realistic human motions without any historical observations. While earlier works~\cite{ormoneit2005representing,urtasun2007modeling} focused on simple, cyclic movements, e.g., using Principal Component Analysis~\cite{ormoneit2005representing} or the Gaussian process latent variable model~\cite{urtasun2007modeling}, 
\begin{figure}[!ht]
    \centering
    \includegraphics[width=\linewidth]{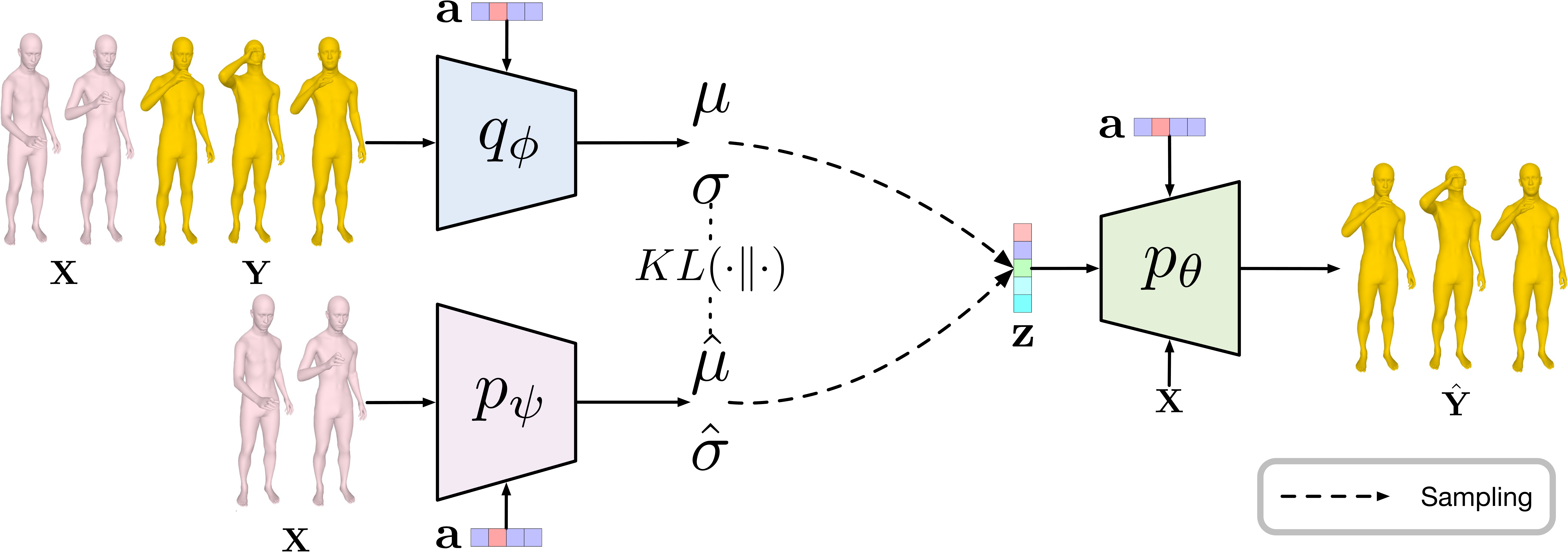} 
    \vspace{-0.5cm}
    \caption{{\bf Overview of our approach.} 
    }
    \label{fig:overview}
    \vspace{-0.5cm}
\end{figure}
recent deep-learning-based methods~\cite{lin2018human,shlizerman2018audio,ahuja2019language2pose,lee2019dancing,li2020learning,guo2020action2motion,li2021learn,petrovich21actor} can handle more complicated motions. In this context, several works have proposed to condition the generated motions on some  auxiliary signal, such as audio/music~\cite{shlizerman2018audio,lee2019dancing,li2020learning,li2021learn} or textual descriptions~\cite{lin2018human,ahuja2019language2pose}. 

The methods most closely related to ours are~\cite{guo2020action2motion,petrovich21actor} which aim to generate action-specific human motions. In particular,~\cite{guo2020action2motion} introduces a frame-level VAE-based model conditioned on the action label, and~\cite{petrovich21actor} a Transformer~\cite{vaswani2017attention}-based VAE model with a sequence-level latent embedding. However, these models can only generate motions depicting individual actions. In principle, given supervised data covering all possible action transitions, they could be trained to generate more complex motions. However, such data cannot practically be obtained. We overcome this by designing a weakly-supervised training strategy that lets us leverage limited, single-action sequences.

\noindent\textbf{Variable-length Motion Prediction.} Although generating sequences of variable length has been well-studied for machine translation~\cite{vaswani2017attention}, it is rarely considered in human motion prediction/synthesis. However, as studied in~\cite{abu2018will} in the context of predicting future actions' semantics and duration from video data, different action categories, or even instances of the same action, vary significantly in length.
Our method produces a variable-length future motion given an action label and a past motion. While~\cite{petrovich21actor} also generates motions of variable lengths, these length must be set manually. In contrast, we automatically find the appropriate duration by learning the distribution of motion lengths.

\noindent\textbf{Action-conditioned Video Generation.} The work most closely related to ours is PSGAN~\cite{yang2018pose}, which aims to predict future 2D human poses given one input image and a target action label. However, with only one input image, PSGAN cannot predict action transitions. Other action-conditioned generative methods include~\cite{wang2020imaginator} and~\cite{kim2020learning}. However, these works aim to generate face images conditioned on emotions, and  the next game screen conditioned on keyboard actions, respectively, which both fundamentally differ from our task.
\section{Our Approach}
Let us now introduce our approach to action-driven stochastic human motion prediction. To represent a human in 3D, we adopt the SMPL model~\cite{SMPL:2015}, which parametrizes a 3D human mesh in terms of shape and pose. 
Since we focus on human motion and not human identity, our model follows that of ACTOR~\cite{petrovich21actor} to only predict the pose parameters. The shape parameters are used for visualization only.
Given an action label represented by a one-hot vector $\mathbf{a}$ and a sequence of $N$ past human poses represented by
$\mathbf{X}=[\mathbf{x}_1,\mathbf{x}_2,\cdots,\mathbf{x}_N]\in \mathbb{R}^{K\times N}$, where $\mathbf{x}_i\in
\mathbb{R}^{K}$ is the pose in the $i$-th frame, our goal is to predict a future motion $\hat{\mathbf{Y}}=[\hat{\mathbf{y}}_{1},\hat{\mathbf{y}}_{2},\cdots,\hat{\mathbf{y}}_{T}]\in \mathbb{R}^{K\times T}$, with $\hat{\mathbf{y}}_i\in
\mathbb{R}^{K}$, representative of
the given action label. To learn to predict transitions between different actions, as discussed in more detail below, we will train our model using data where $\mathbf{X}$ and the corresponding ground-truth future motion $\mathbf{Y}$ depict different actions. This will eventually allow us to predict future motions for sequences of action labels, by recursively treating the previous prediction as historical information.

\subsection{Action-driven Stochastic Motion Prediction}
To predict action-driven future motions, we design a model based on conditional VAEs (CVAEs)~\cite{kingma2013auto}, whose goal is to model the
conditional distribution $p(\mathbf{Y}|\mathbf{X},\mathbf{a})$.
Specifically, as shown in Fig.~\ref{fig:overview}, we first model the posterior distribution
$q_{\phi}(\mathbf{z}|\mathbf{Y},\mathbf{X},\mathbf{a})$ via a neural network, the encoder,
where $\mathbf{z}$ is a latent random variable, and $\phi$ denotes the parameters of the encoder. From
the latent variable $\mathbf{z}$, the CVAE then aims to reconstruct the future motion
$\mathbf{Y}$ using another neural network, the decoder, expressed as $p_{\theta}(\mathbf{Y}|\mathbf{z},\mathbf{X},\mathbf{a})$, with parameters $\theta$. The evidence lower bound (ELBO) of the conditional distribution
$p(\mathbf{Y}|\mathbf{X},\mathbf{a})$ can then be written as
\begin{align}
    \log p(\mathbf{Y}|\mathbf{X},\mathbf{a}) \geq  &\mathbb{E}_{q_{\phi}(\mathbf{z}|\mathbf{Y},\mathbf{X},\mathbf{a})}[\log p_{\theta}(\mathbf{Y}|\mathbf{z},\mathbf{X},\mathbf{a})] - \nonumber\\
    &KL(q_{\phi}(\mathbf{z}|\mathbf{Y},\mathbf{X},\mathbf{a})\|p_{\psi}(\mathbf{z}|\mathbf{X},\mathbf{a}))\;,
\end{align}
where $p_{\psi}(\mathbf{z}|\mathbf{X},\mathbf{a})$ is the prior distribution of the latent variable $\mathbf{z}$, modeled by a neural network with parameters $\psi$, and $KL(\cdot||\cdot)$ is the KL divergence between two distributions. Training the CVAE then aims to maximize the log probability $\log p(\mathbf{Y}|\mathbf{X},\mathbf{a})$ by maximizing the ELBO.

In practice, the KL divergence term in the ELBO can be computed as,
\vspace{-0.3cm}
\begin{align}
    \mathcal{L}_{\text{KL}} &= KL(\mathcal{N}(\mathbf{\mu},\mathrm{diag}(\mathbf{\sigma}^2))\|\mathcal{N}(\hat{\mathbf{\mu}},\mathrm{diag}(\hat{\mathbf{\sigma}}^2))) \nonumber \\ 
    &=\frac{1}{2}\sum_{i=1}^{D}\left(\log\frac{\hat{\mathbf{\sigma}_i}^2}{\mathbf{\sigma}_i^2}+\frac{\mathbf{\sigma}_i^2+(\mathbf{\mu}_i-\hat{\mathbf{\mu}}_i)^2}{\hat{\mathbb{\sigma}}_i^2}-1\right)\;,
\end{align}\\[-0.3cm]
where $\mathcal{N}(\mathbf{\mu},\mathrm{diag}(\mathbf{\sigma}^2))$ and $\mathcal{N}(\hat{\mathbf{\mu}},\mathrm{diag}(\hat{\mathbf{\sigma}}^2))$ are the posterior and prior distributions, whose means and standard deviations are produced by the encoder $q_{\phi}$ and  the prior network $p_{\psi}$, respectively, and $D$ is the dimension of $\mathbf{z}$. 

\begin{figure*}[!ht]
\vspace{-0.3cm}
    \centering
    \includegraphics[width=0.8\textwidth]{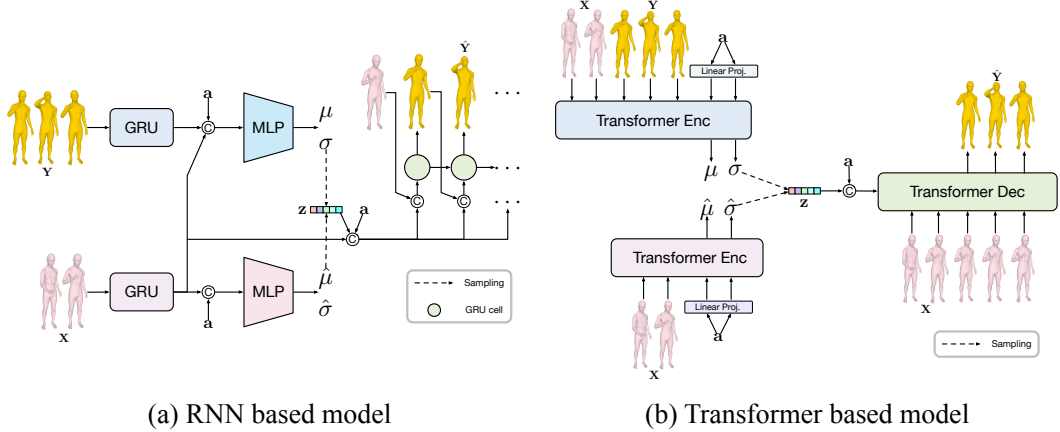} 
    \vspace{-0.3cm}
    \caption{{\bf Network structure.} We explore the use of two different temporal encoding structures to build our VAE: RNNs and Transformers.}
    \label{fig:networks}
    \vspace{-0.4cm}
\end{figure*}
During training, the random variable $\mathbf{z}$ is sampled from the posterior distribution via the reparameterization trick~\cite{kingma2013auto}, i.e., $\mathbf{z} = \mathbf{\epsilon}\odot\mathbf{\sigma} + \mathbf{\mu}$, where $\mathbf{\epsilon}\sim\mathcal{N}(\mathbf{0},\mathbf{I})$. Given $\mathbf{z}$, the past poses $\mathbf{X}$ and the action label $\mathbf{a}$, the goal of decoder $p_{\theta}$ is to reconstruct the true future motion. This lets us express the (negative of the) first term of the ELBO as the reconstruction loss
\vspace{-0.3cm}
\begin{align}
    \mathcal{L}_{\text{rec}}=\frac{1}{T}\sum_{i=1}^{T}\|\hat{\mathbf{y}}_i-\mathbf{y}_i\|_{2}^2\;,\label{eq:rec1}
\end{align}\\[-0.3cm]
where $\hat{\mathbf{Y}}=[\hat{\mathbf{y}}_{1},\hat{\mathbf{y}}_{2},\cdots,\hat{\mathbf{y}}_{T}]$ is the future motion generated by the decoder.

Note that, during training, sampling ${\bf z}$ involves the encoder $q_{\phi}$, which relies on the ground-truth motion ${\bf Y}$. Since, at test time, the ground-truth future motion is unknown, we sample the random variable from the prior distribution.

\subsection{Weakly-supervised Transitions Learning}
Natural human movement involves transitions between different action categories. The ability to generate these transitions is therefore critical for the success and realism of human motion modeling methods. However, acquiring training data that covers all possible action transitions is virtually intractable, and thus existing human motion datasets typically contain motions depicting individual actions only, without any transitions. To nonetheless effectively leverage this data to learn action transitions, we create synthetic motions by combining historical motions from one action category with future motions from another. As these synthetic motions still do not contain realistic transitions, we introduce a weakly-supervised training strategy to learn to generate plausible transitions.

More specifically, given a historical motion $\mathbf{X}=[\mathbf{x}_1,\mathbf{x}_2,\cdots,\mathbf{x}_N]$ from one action, we take motion $\mathbf{Y}'=[\mathbf{y}'_{1},\mathbf{y}'_{2},\cdots,\mathbf{y}'_{T}]$ from another action to be the continuation of $\mathbf{X}$ after $T_0$ frames. However, both the number of frames $T_0$  and poses in these frames are unknown, and one cannot assume $T_0$ to be constant for any pair of historical and future motions. To address this, we define $T_0$ to be a function of the last pose of $\mathbf{X}$ and of the first one of $\mathbf{Y}'$, i.e., $T_0 = f(\mathbf{x}_N,\mathbf{y}'_{1})$,
where $f:\mathbb{R}^{K}\times\mathbb{R}^{K} \rightarrow \mathbb{N}$. In practice, we found that a simple linear function suffices, and thus write $T_0 = \lfloor {k\|\mathbf{x}_N-\mathbf{y}'_{1}\|_2}\rfloor$, where $k>0$ is computed from the training data. Details about computing $k$ are provided in the supplementary material.

To account for the fact that the poses within the transition sequence, namely poses for $T_0$ frames, are unknown, we leverage a simple temporal smoothness prior based on the intuition that the transition from $\mathbf{X}$ to $\mathbf{Y}'$ should form a smooth sequence. Inspired by~\cite{akhter2009nonrigid,mao2019learning,huang2017towards}, we make use of the Discrete Cosine Transform (DCT) to define our smoothness prior, exploiting the insight that a smooth trajectory can be accurately represented by low frequency DCT bases. More precisely, let $\hat{\mathbf{Y}}=[\hat{\mathbf{y}}_{1},\hat{\mathbf{y}}_{2},\cdots,\hat{\mathbf{y}}_{T_0+T}]$ denote the prediction of our model. We first concatenate the last $L$ poses of the history and the first $L$ ones of the prediction to form a sequence of length $2L$ denoted by $\hat{\mathbf{Z}}=[\mathbf{x}_{N-L+1},\mathbf{x}_{N-L+2},\cdots,\mathbf{x}_N,\hat{\mathbf{y}}_{1},\hat{\mathbf{y}}_{2},\cdots,\hat{\mathbf{y}}_{L}]$, where $L\leq N$ and $\hat{\mathbf{Z}}\in\mathbb{R}^{K\times 2L}$. We then approximate this sequence with the first $M$ DCT bases as $\Tilde{\mathbf{Z}} = \hat{\mathbf{Z}}\mathbf{D}\mathbf{D}^T$,
where $\mathbf{D}\in\mathbb{R}^{2L\times M}$ encodes the low-frequency DCT basis and $M\leq 2L$. Given $\hat{\mathbf{Z}}$ and its approximation $\Tilde{\mathbf{Z}}$, we define our temporal smoothness prior as the loss
\vspace{-0.3cm}
\begin{align}
    \mathcal{L}_{\text{\text{smooth}}} = \frac{1}{2L}\sum_{i=1}^{2L} \|\hat{\mathbf{z}}_i-\Tilde{\mathbf{z}}_i\|_2^2\;,
\end{align}\\[-0.3cm]
where $\hat{\mathbf{z}}_i$ and $\Tilde{\mathbf{z}}_i$ are the $i$-th pose in $\hat{\mathbf{Z}}$ and $\Tilde{\mathbf{Z}}$, respectively. 

Since we only have ground-truth supervision for the last $T$ predicted frames, we redefine the reconstruction loss as
\vspace{-0.3cm}
\begin{align}
    \mathcal{L}_{\text{rec}}=\frac{1}{T}\sum_{i=1}^{T}\|\hat{\mathbf{y}}_{T_0+i}-\mathbf{y}_i\|_2^2\;.
\end{align}\\[-0.4cm]
Note that our formulation still allows us to exploit data where $\mathbf{Y}'$ and $\mathbf{X}$ are from the same motion sequence by simply setting the corresponding $T_0$ to zero.

Altogether, we express our complete training loss as
\vspace{-0.3cm}
\begin{align}
    \mathcal{L} = \lambda_{\text{rec}}\mathcal{L}_{\text{rec}}+\lambda_{\text{smooth}}\mathcal{L}_{\text{smooth}}+\mathcal{L}_{\text{KL}}\;,
\end{align}\\[-0.6cm]
where $\lambda_{\text{rec}}$ and $\lambda_{\text{max}}$ are hyper-parameters setting the relative influence of the different terms.
\vspace{-0.2cm}
\subsection{Variable-length Motion Prediction}
Generating sequences of variable length has been well-studied in the field of Natural Language Processing (NLP)~\cite{vaswani2017attention}, where the standard strategy consists of predicting a specific stop token. Here, instead of predicting a stop token, which is ill-defined for human motion, we simply encourage the model to generate static poses (the last pose of the ground-truth motion) after reaching the motion end during training. Specifically, we make the model generate $P$ additional frames, leading to a future sequence of $T_0+T+P$ frames ($\hat{\mathbf{Y}}=[\hat{\mathbf{y}}_{1},\hat{\mathbf{y}}_{2},\cdots,\hat{\mathbf{y}}_{T+T_0+P}]$). We then supervise these additional frames with the last ground-truth future pose. Combining this with the normal supervision of the other frames yields the updated reconstruction loss
\begin{align}
    \mathcal{L}_{\text{rec}}=&\frac{1}{T+P}\sum_{i=1}^{T+P}\|\hat{\mathbf{y}}_{T_0+i}-\mathbf{y}_{i}\|_2^2\;, 
\end{align}

During testing, as we do not know the length of the predicted future, we stop prediction when the variance of the last $Q$ consecutive frames falls below a threshold.
Specifically, given the predicted future motion $\hat{\mathbf{Y}}$, we compute, for $Q$ consecutive frames starting from the $i$-th one,
\vspace{-0.3cm}
\begin{align}
    v_i = \frac{1}{Q}\sum_{j=i}^{i+Q}\|\hat{\mathbf{y}}_j-\frac{1}{Q}\sum_{k=i}^{i+Q}\hat{\mathbf{y}}_k\|_2\;,
\end{align}\\[-0.5cm]
where $i\in[1,2,\cdots,T_{\text{max}}-Q]$, and $T_{\text{max}}$ is the maximum number of frames the model can predict. We stop the prediction at frame $i$ if $v_i < \delta$.

\subsection{Network Structure}
To show the generality of our approach, we exploit it using  two different temporal encoding structures, namely, Recurrent Neural Networks (RNNs) and Transformers~\cite{vaswani2017attention}. 

For our RNN-based model, shown in Fig~\ref{fig:networks}~(a), we build the encoder $q_{\phi}$, the prior $p_{\psi}$ and the decoder $p_{\theta}$ using Gated Recurrent Units (GRUs). In particular, the encoder $q_{\phi}$ first uses GRUs to encode the historical human poses $\mathbf{X}$ and the future human motion $\mathbf{Y}$ to temporal features. These temporal features concatenated with the action token obtained from a fully connected layer are then fed into a fully connected network that predicts the parameters (the mean $\mu$ and the standard deviations $\sigma$) of the posterior distribution. The prior network produces the parameters (the mean $\hat{\mu}$ and the standard deviations $\hat{\sigma}$) of the prior distributions in a similar manner. Given the latent code $\mathbf{z}$ sampled from either the posterior (during training) or the prior (during testing), the temporal features of the historical motion $\mathbf{X}$ and the action label $\mathbf{a}$, the decoder again uses GRUs to predict the future poses in an autoregressive manner.

We show our Transformer-based model in Fig.~\ref{fig:networks}~(b). For the encoder and prior network, we adopt the same strategy as~\cite{petrovich21actor}, which was inspired by BERT~\cite{devlin2018bert} in NLP and ViT~\cite{dosovitskiy2020image} in Computer Vision. In particular, we append two extra tokens obtained from the action label $\mathbf{a}$ to aggregate temporal information to predict the parameters of the posterior and prior distributions. For the decoder, we pad the historical human motion with its last pose, forming a longer sequence, and then input the padded sequence to the Transformer-based decoder that outputs the future motion. To introduce the action information and the latent random code as conditions, we further use the pseudo self attention strategy proposed in~\cite{ziegler2019encoder}.
\section{Experiments}
\subsection{Datasets}\label{sec:dataset}
We evaluate our method on three different datasets. Each motion sequence in these datasets is annotated with a single action label, except for BABEL~\cite{BABEL:CVPR:2021}. Some information for each dataset is provided in Table~\ref{tab:dataset}. We also evaluate on the dataset HumanAct12~\cite{guo2020action2motion}. The results are in the supplementary material.

\vspace{-0.3cm}
\begin{table}[!ht]
    \centering
    \resizebox{\linewidth}{!}{
    \begin{tabular}{c|ccccc}
    \toprule
    Dataset & motion len. & train  & test & transi. & action \\
    \midrule
    GRAB~\cite{GRAB:2020,Brahmbhatt_2019_CVPR} & 100-501 & 1149 & 319 & 0 & 4\\
    NTU RGBD~\cite{shahroudy2016ntu,liu2019ntu} & 35-201 & 3399 & 361 & 0 & 13 \\
    BABEL~\cite{BABEL:CVPR:2021} & 30-300 & 9643 & 3477 & 2584 & 20\\
    \bottomrule
    \end{tabular}
    }
    \vspace{-0.3 cm}
    \caption{\textbf{Datasets' details.} We list the range of motion length in frames, the number of training/testing samples, the number of training samples with action transitions and the number of actions in each dataset.}\label{tab:dataset}
    \vspace{-0.3 cm}
\end{table}
\noindent\textbf{GRAB}~\cite{GRAB:2020,Brahmbhatt_2019_CVPR} consists of 10 subjects interacting with 51 different objects, performing 29 different actions. Since, for most actions, the number of samples is too small for training, we 
choose the four action categories with the most motion samples, i.e., Pass, Lift, Inspect and Drink. We use 8 subjects (S1-S6, S9, S10) for training and the remaining 2 subjects (S7, S8) for testing. In all cases, we remove the global translation. The original frame rate is 120 Hz. To further enlarge the size of the dataset, we downsample the sequences to 15-30 Hz.
Our model is trained to observe 25 frames to predict the future. The observed frames and the future ones are from either the same or different motions.

\noindent\textbf{NTU RGB-D}~\cite{shahroudy2016ntu,liu2019ntu} (NTU). We use the subset of 13 actions of~\cite{guo2020action2motion}, with noisy SMPL parameters estimated by VIBE~\cite{kocabas2020vibe}. As for GRAB, we remove the global translation. While~\cite{guo2020action2motion} used all the data for training, we split the dataset into training and testing by subjects.  
Our model is trained to observe 10 past frames.

\noindent\textbf{BABEL}~\cite{BABEL:CVPR:2021} is a subset of the AMASS dataset~\cite{AMASS:2019} with per-frame action annotations. Since there are multiple action labels in one motion sequence, we split the dataset into two parts: single-action sequences and sequences that depict transitions between two actions. We downsample all motion sequences to 30 Hz. For single-action motions, we first divide the long motions into several short ones. Each short motion performs one single action. 
and the remove sequences that are too short ($<1$ second). We also eliminate the action labels with too few samples ($<60$) or overlap with other actions, e.g. foot movement sequences sometimes overlap with kicking. This leaves us with 20 action labels. We complement this data with the sequences with transitions that contain these 20 actions. During training, our model observes 10 past frames to predict the future.
\subsection{Evaluation Metrics and Baseline}\label{sec:eval}
\textbf{Metrics.} We follow the similar evaluation protocol as for human motion synthesis/prediction~\cite{guo2020action2motion,petrovich21actor,yuan2020dlow} and employ the following metrics to evaluate our method. 

(1) To measure the distribution similarity between the generated sequences and the ground-truth motions, we adopt the Fréchet Inception Distance (FID)~\cite{heusel2017gans}
\vspace{-0.3cm}
\begin{align}
    FID = &\|\mu_{\text{gen}} - \mu_{\text{gt}}\|^2 \nonumber\\ 
    &+ Tr(\Sigma_{\text{gen}} + \Sigma_{\text{gt}} - 2 (\Sigma_{\text{gen}}\Sigma_{\text{gt}})^{1/2})\;,
\end{align}\\[-0.5cm]
where $\mu_{\cdot}\in \mathbb{R}^{F}$ and $\Sigma_{\cdot}\in \mathbb{R}^{F\times F}$ are the mean and covariance matrix of perception features obtained from a pretrained action recognition model, with $F$ the dimension of the perception features. The detail of the action recognition model is included in the supplementary material. $Tr(\cdot)$ computes the trace of a matrix. 

(2) To evaluate motion realism, we report the action recognition accuracy of the generated motions using the same pretrained action recognition model as above. 

(3) To evaluate per-action diversity, we measure the pairwise distance between the multiple future motions generated from the same historical motion and action label\footnote{Note that we only report diversity per action because motions of different actions are inherently diverse.}. Specifically, given a set of future motions $\{\hat{\mathbf{Y}}^i\}_{i=1}^{S}$ predicted by our model, the diversity is computed as
\vspace{-0.3cm}
\begin{align}
\resizebox{0.88\linewidth}{!}{$
    Div = \frac{2}{S(S-1)}\sum_{i=1}^{S}\sum_{j=i+1}^{S}\frac{1}{T_{\text{max}}} \sum_{k=1}^{T_{\text{max}}} \|\hat{\mathbf{y}}^{i}_{k}-\hat{\mathbf{y}}^{j}_{k}\|_2 \;,
$}
\end{align}\\[-0.5cm]
where $T_{\text{max}}$ is the maximum number of frames our model can predict, and $\hat{\mathbf{y}}^{i}_{k}$ represents the $k$-th frame of motion $\hat{\mathbf{Y}}^{i}$.

To calculate above mentioned diversity, we assume that the model generates the maximum number of future frames in all cases. To further evaluate the diversity across variable length future motions, we compute the average per-action diversity after performing Dynamic Time Warping (DTW)~\cite{zhang2019predicting} 
With a minor abuse of notation, let $\{\hat{\mathbf{Y}}^i\}_{i=1}^{S}$ denote the set of variable length predictions. DTW then temporally aligns any pair of motions as $\Tilde{\mathbf{Y}}^i,\Tilde{\mathbf{Y}}^j = DTW(\hat{\mathbf{Y}}^i,\hat{\mathbf{Y}}^j)\;,$
where $\Tilde{\mathbf{Y}}^i$ and $\Tilde{\mathbf{Y}}^j \in\mathbb{R}^{K\times T_{i,j}}$ have the same number of frames ($T_{i,j}$). We then compute the diversity after DTW as
\vspace{-0.3cm}
\begin{align}
\resizebox{0.85\linewidth}{!}{$
    Div_{\text{w}} = \frac{2}{S(S-1)}\sum_{i=1}^{S}\sum_{j=i+1}^{S}\frac{1}{T_{i,j}} \sum_{k=1}^{T_{i,j}} \|\Tilde{\mathbf{y}}^{i}_{k}-\Tilde{\mathbf{y}}^{j}_{k}\|_2 \;.
$}
\end{align}\\[-1cm]

(4) To measure the prediction accuracy, we adopt the Average Displacement Error (ADE) computed as
\vspace{-0.3cm}
\begin{align}\resizebox{0.55\linewidth}{!}{$
ADE = \min\limits_{i} \frac{1}{T} \sum_{k=1}^{T} \|\hat{\mathbf{y}}^{i}_{k}-\mathbf{y}_{k}\|_2 \;,
$}
\vspace{-0.3cm}
\end{align}\\[-0.5cm]
where $T$ is the length of the ground-truth future motion, $\hat{\mathbf{y}}^{i}_{k}$ is the $k$-th frame of the $i$-th sample generated by the model with the ground-truth action label\footnote{Since we only have the ground-truth future motion for the ground-truth action label.} and $\mathbf{y}_{k}$ the corresponding ground truth. Similarly to the diversity, we report the ADE after DTW ($ADE_{\text{w}}$).

\textbf{Baselines.} Since there is no prior work that tackles the task we
introduce, we adapt the state-of-the-art action-specific human motion
synthesis methods, Action2Motion~\cite{guo2020action2motion},
ACTOR~\cite{petrovich21actor}, and stochastic human motion prediction method, DLow~\cite{yuan2020dlow}, to our task. Action2Motion~\cite{guo2020action2motion} relies on a frame-wise motion VAE with GRUs to encode the temporal information. We adapt their VAE so as to take the temporal feature of historical poses as an additional input for both encoding and decoding. This temporal feature is extracted from a GRU-based temporal data encoding module. Similarly, we modify the transformer decoder of ACTOR~\cite{petrovich21actor} to condition it on the historical motion. Furthermore, we adapt the VAE in DLow~\cite{yuan2020dlow} to take the action label as input.

\textbf{Implementation details.}
 We implement our models in Pytorch~\cite{paszke2017automatic} and train them using the ADAM~\cite{kingma2014adam} optimizer for 500 epochs. 
 We use different hyperparameters for different models. In particular, for RNN-based model, the initial learning rate is 0.001 on BABEL and 0.002 on all the other datasets. We set the loss weights ($\lambda_{\text{rec}}$, $\lambda_{\text{smooth}}$)  to (50.0, 10.0) for BABEL dataset and (100.0,100.0) for all the other ones. For Transformer-based model, the initial learning rate is 0.0001 on BABEL and 0.0005 on all the other datasets. The loss weights ($\lambda_{\text{rec}}$, $\lambda_{\text{smooth}}$) are set to (100.0, 10.0) for BABEL dataset and (1000.0,100.0) for all the other ones. Additional details are in the supplementary material.

\subsection{Results}
\textbf{Quantitative results.} In Table~\ref{tab:quantitative}, 
we compare our results with those of baselines on GRAB, NTU RGB-D and BABEL. Given a past motion, all models predict multiple future motions conditioned on any given action label.
Our approach, based on either RNNs or Transformers, outperforms the baselines on almost all metrics. In general, the RNN-based model performs better than the Transformer-based one. We expect this to be due to the datasets being too small to train the Transformer-based model from scratch. 

\begin{figure*}[!ht]
    \vspace{-0.3cm}
    \centering
    \includegraphics[width=0.7\textwidth]{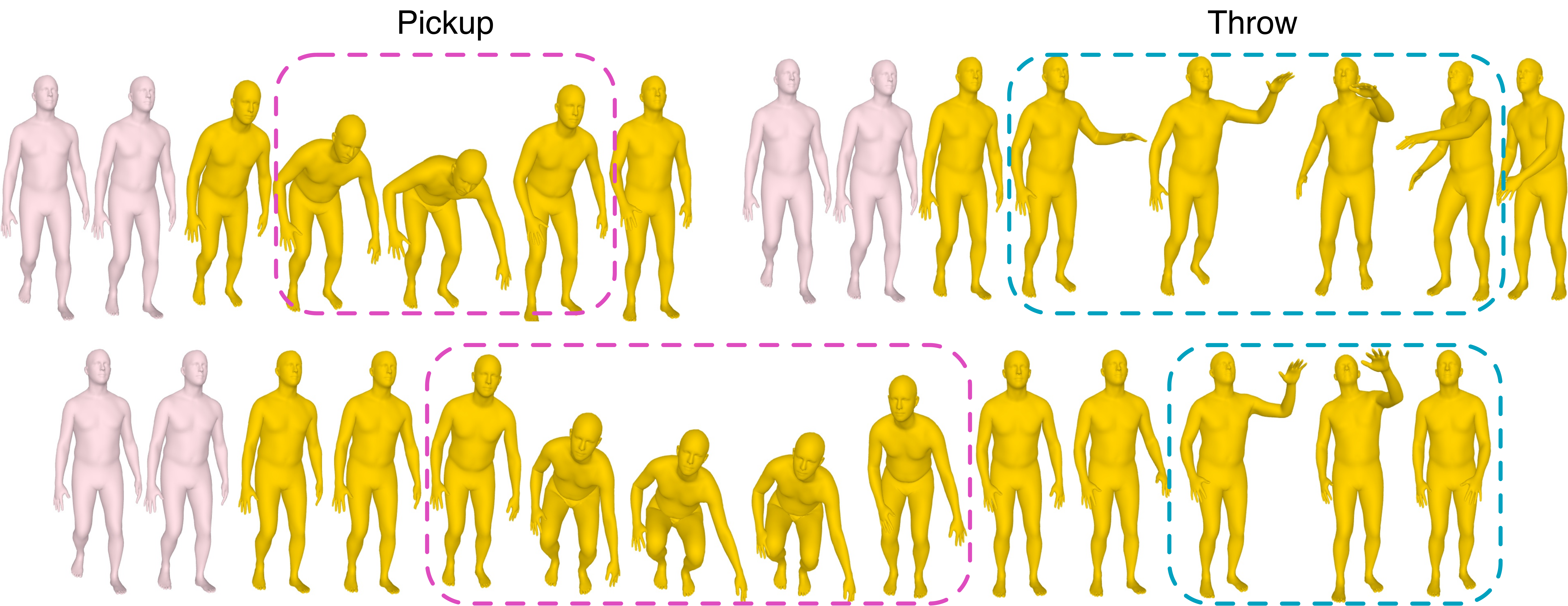} 
    \vspace{-0.3cm}
    \caption{\textbf{Results on NTU RGB-D.} Given the same history (pink), our model can generate future motions (yellow) of different actions, e.g., ``Pickup'' (top left) or ``Throw'' (top right). Moreover, it can also generate motions depicting a sequence of multiple actions (bottom).}
    \label{fig:ntu_qual}
    \vspace{-0.4cm}
\end{figure*}
\begin{table}[!ht]
\vspace{-0.2cm}
    \centering
    \resizebox{\linewidth}{!}{
    \begin{tabular}{cc|ccccc}
    \toprule
    &Method & Acc$\uparrow$ & $FID_{\text{tr}}\downarrow$ & $FID_{\text{te}}\downarrow$ & $Div_{\text{w}}\uparrow$  & $Div\uparrow$\\
    \midrule
    \multirow{5}{*}{\rotatebox{90}{GRAB}}
    &Act2Mot~\cite{guo2020action2motion} & $70.6^{\pm 1.3}$ & $80.22^{\pm 6.64}$ & $47.81^{\tiny{\pm 1.09}}$ & $0.50^{\pm 0.00}$ & $0.76^{\pm 0.01}$ \\
     &DLow~\cite{yuan2020dlow} & $67.6^{\pm 0.7}$ & $127.49^{\pm 6.90}$ & $\textbf{22.71}^{\pm 2.79}$ & $0.74^{\pm 0.01}$ & $0.92^{\pm 0.01}$\\
    &ACTOR~\cite{petrovich21actor} & $83.0^{\pm 0.3}$ & $62.68^{\pm 1.26}$ & $114.85^{\pm 3.46}$ & $1.06^{\pm 0.00}$ & $1.04^{\pm 0.00}$ \\\cmidrule{2-7}
    &Ours (RNN) & $\textbf{92.6}^{\pm 0.6}$ & $\textbf{44.59}^{\pm 1.39}$ & $38.03^{\pm 1.49}$ & $\textbf{1.10}^{\pm 0.01}$ & $\textbf{1.37}^{\pm 0.01}$ \\
    &Ours (Tran.) & $85.5^{\pm 1.2}$ & $48.58^{3.05}$ & $25.72^{\pm 2.16}$ & $1.05^{\pm 0.01}$ & $1.08^{\pm 0.01}$\\
    \bottomrule
    \multirow{5}{*}{\rotatebox{90}{NTU}}
    &Act2Mot~\cite{guo2020action2motion} & $66.3^{\pm 0.2}$ & $144.98^{\pm 2.44}$ & $113.61^{\pm 0.84}$ & $0.75^{\pm 0.01}$ & $1.19^{\pm 0.01}$ \\
    &DLow~\cite{yuan2020dlow} & $70.6^{\pm 0.2}$ & $151.11^{\pm 1.25}$ & $157.54^{\pm 1.62}$ & $0.97^{\pm 0.00}$ & $1.21^{\pm 0.00}$ \\
    &ACTOR~\cite{petrovich21actor} & $66.3^{\pm 0.1}$ & $355.69^{\pm 5.74}$ & $193.58^{\pm 2.91}$ & $\textbf{1.84}^{\pm 0.00}$ & $2.07^{\pm 0.00}$ \\\cmidrule{2-7}
    &Ours (RNN) & $\textbf{76.0}^{\pm 0.2}$ & $\textbf{72.18}^{\pm 0.93}$  & $\textbf{111.01}^{\pm 1.28}$ & $1.25^{\pm 0.00}$ & $\textbf{2.20}^{\pm 0.00}$ \\
    &Ours (Tran.) & $71.3^{\pm 0.2}$ & $83.14^{\pm 1.74}$  & $114.62^{\pm 0.93}$ & $1.25^{\pm 0.00}$ & $2.19^{\pm 0.01}$ \\
    \bottomrule
    \multirow{5}{*}{\rotatebox{90}{BABEL}}
    &Act2Mot~\cite{guo2020action2motion} & $14.8^{\pm 0.2}$ & $42.02^{\pm 0.40}$ & $37.41^{\pm 0.47}$ & $0.79^{\pm 0.01}$ & $1.10^{\pm 0.01}$ \\
    &DLow~\cite{yuan2020dlow} & $12.7^{\pm 0.2}$ & $27.99^{\pm 0.45}$ & $24.18^{\pm 0.59}$ & $0.65^{\pm 0.00}$ & $0.90^{\pm 0.00}$ \\
    &ACTOR~\cite{petrovich21actor} & $40.9^{\pm 0.2}$ & $29.34^{\pm 0.10}$ & $30.31^{\pm 0.16}$ & $\textbf{2.94}^{\pm 0.00}$ & $\textbf{2.71}^{\pm 0.00}$ \\\cmidrule{2-7}
    &Ours (RNN) & $\textbf{49.6}^{\pm 0.4}$ & $22.54^{\pm 0.27}$ & $22.39^{\pm 0.36}$ & $1.35^{\pm 0.00}$ & $1.74^{\pm 0.00}$ \\
    &Ours (Tran.) & $39.5^{\pm 0.3}$ & $\textbf{20.02}^{\pm 0.24}$ & $\textbf{19.41}^{\pm 0.35}$ & $1.39^{\pm 0.00}$ & $1.82^{\pm 0.01}$ \\
    \bottomrule
    \end{tabular}
    }
    \vspace{-0.3 cm}
    \caption{\textbf{Quantitative results.} 
    We report the action recognition accuracy (Acc), the FID to training data ($FID_{\text{tr}}$) and to the testing split ($FID_{\text{te}}$), and the diversity before ($Div$) and after DTW ($Div_{\text{w}}$). We adapt Action2Motion~\cite{guo2020action2motion}, ACTOR~\cite{petrovich21actor} and DLow~\cite{yuan2020dlow} to our task.
    }\label{tab:quantitative}
    \vspace{-0.7 cm}
\end{table}
In Table~\ref{tab:ade}, we compare the prediction accuracy ($ADE$, $ADE_{\text{w}}$) of our results with the baselines. Here, for each past motion, each model predicts multiple future motions using the ground-truth action label. The prediction accuracy (ADE) is then computed based on the future motion yielding the minimum error. Because our model is trained to predict not only the ground-truth future but also motions with different action labels, it may sacrifice some accuracy when evaluated on the ground-truth future only, as on NTU and BABEL.

During training, our model only takes one action label and a motion history as input. During testing, to predict future motions for a sequence of action labels of arbitrary length, we follow a recursive strategy.
We evaluate this in the case of 5-action sequences. Specifically, we randomly sampled sequences of 5 action labels to generate future motions in an autoregressive manner, and report the results at each prediction step, i.e., corresponding to each action label. The results shown in Table~\ref{tab:actseq} indicate that our model remains stable. Note that the performance gap between the 1st and 2nd step on NTU may be caused by the fact that our model is trained with the jittery ``ground-truth'' NTU motion history, while at the 2nd step, it starts taking as input the smooth motion predicted during the 1st step.

\textbf{Qualitative results.} In Fig.~\ref{fig:opening}, we show diverse futures generated by our model on GRAB given the same past motion and the same action sequence. Additional qualitative results on the NTU RGB-D dataset are provided in Fig.~\ref{fig:ntu_qual}. Given the same historical poses, our model can generate futures of different actions and sequences of multiple actions. More results are provided in the supplementary material.

\begin{table}[!ht]
    \vspace{-0.2 cm}
    \centering
    \resizebox{\linewidth}{!}{
    \begin{tabular}{c|cc|cc|cc}
    \toprule
    & \multicolumn{2}{c|}{GRAB} & \multicolumn{2}{c|}{NTU}  & \multicolumn{2}{c}{BABEL} \\
    Method & $ADE_{\text{w}}\downarrow$  & $ADE\downarrow$ & $ADE_{\text{w}}\downarrow$  & $ADE\downarrow$ & $ADE_{\text{w}}\downarrow$  & $ADE\downarrow$ \\
    \midrule
    Act2Mot~\cite{guo2020action2motion} & $1.92^{\pm 0.03}$ & $2.28^{\pm 0.03}$ & $\textbf{0.78}^{\pm 0.01}$ & $\textbf{1.11}^{\pm 0.01}$ & $1.25^{\pm 0.02}$ & $1.27^{\pm 0.01}$ \\
    DLow~\cite{yuan2020dlow} & $1.78^{\pm 0.03}$ & $1.96^{\pm 0.03}$ & $0.95^{\pm 0.01}$ & $1.20^{\pm 0.01}$ & $\textbf{1.10}^{\pm 0.01}$ & $\textbf{1.19}^{\pm 0.01}$ \\
    ACTOR~\cite{petrovich21actor} & $2.41^{\pm 0.02}$ & $2.57^{\pm 0.02}$ & $1.26^{\pm 0.01}$ & $1.49^{\pm 0.01}$ & $2.19^{\pm 0.02}$ & $2.29^{\pm 0.02}$ \\\midrule
    Ours (RNN) & $1.73^{\pm 0.02}$ & $\textbf{1.93}^{\pm 0.03}$ & $0.89^{\pm 0.01}$ & $1.20^{\pm 0.01}$ & $1.31^{\pm 0.00}$ & $1.47^{\pm 0.01}$ \\
    Ours (Tran.) &  $\textbf{1.69}^{\pm 0.02}$ & $\textbf{1.93}^{\pm 0.03}$ & $0.84^{\pm 0.01}$ & $1.23^{\pm 0.01}$ & $1.24^{\pm 0.01}$ & $1.40^{\pm 0.02}$ \\
    \bottomrule
    \end{tabular}
    }
    \vspace{-0.3 cm}
    \caption{\textbf{Results of prediction accuracy} Our model may trade some performance with GT action label for the ability of predicting future motion of different action labels.}\label{tab:ade}
    \vspace{-0.2 cm}
\end{table}
\begin{table}[!ht]
    \vspace{-0.3 cm}
    \centering
    \resizebox{\linewidth}{!}{
    \begin{tabular}{cc|ccccc}
    \toprule
    & & \multicolumn{5}{c}{Prediction Step} \\
    & Metrics & $1_{st}$  & $2_{nd}$ & $3_{rd}$  & $4_{th}$ & $5_{th}$ \\
    \midrule
    \multirow{5}{*}{\rotatebox{90}{GRAB\;\;\;\;}} 
    & Acc$\uparrow$ & $92.6^{\pm 0.6}$ & $94.3^{\pm 0.6}$ & $93.4^{\pm 0.9}$ & $93.5^{\pm 0.6}$ & $92.6^{\pm 1.0}$ \\
    & $FID_{\text{tr}}\downarrow$ & $44.59^{\pm 1.39}$ & $31.45^{\pm 6.73}$ & $31.53^{\pm 6.36}$ & $38.92^{\pm 6.31}$ & $43.14^{\pm 10.25}$ \\
    & $FID_{\text{te}}\downarrow$ & $38.03^{\pm 1.49}$ & $74.85^{\pm 13.84}$ & $91.65^{\pm 7.21}$ & $111.36^{\pm 24.36}$ & $117.30^{\pm 13.99}$ \\
    & $Div_{\text{w}}\uparrow$ & $1.10^{\pm 0.01}$ & $1.31^{\pm 0.01}$ & $1.33^{\pm 0.01}$ & $1.32^{\pm 0.03}$ & $1.34^{\pm 0.01}$ \\
    & $Div\uparrow$ & $1.37^{\pm 0.01}$ & $1.60^{\pm 0.01}$ & $1.62^{\pm 0.02}$ & $1.61^{\pm 0.03}$ & $1.64^{\pm 0.02}$ \\ \midrule
    \multirow{5}{*}{\rotatebox{90}{NTU\;\;}} 
    & Acc$\uparrow$ & $76.0^{\pm 0.2}$ & $61.9^{\pm 0.7}$ & $61.4^{\pm 0.7}$ & $60.6^{\pm 0.6}$ & $60.1^{\pm 0.6}$ \\
    & $FID_{\text{tr}}\downarrow$ & $72.18^{\pm 0.93}$ & $219.08^{\pm 13.68}$ & $248.21^{\pm 13.65}$ & $243.57^{\pm 7.11}$ & $240.40^{\pm 11.43}$ \\
    & $FID_{\text{te}}\downarrow$ & $111.01^{\pm 1.28}$ & $286.82^{\pm 10.15}$ & $334.42^{\pm 16.71}$ & $334.94^{\pm 4.53}$ & $316.87^{\pm 13.28}$ \\
    & $Div_{\text{w}}\uparrow$ & $1.25^{\pm 0.00}$ & $1.22^{\pm 0.02}$ & $1.23^{\pm 0.01}$ & $1.22^{\pm 0.02}$ & $1.21^{\pm 0.01}$ \\
    & $Div\uparrow$ & $2.20^{\pm 0.00}$ & $2.16^{\pm 0.04}$ & $2.18^{\pm 0.01}$ & $2.17^{\pm 0.04}$ & $2.15^{\pm 0.02}$ \\ \midrule
    \multirow{5}{*}{\rotatebox{90}{BABEL\;\;}} 
    & Acc$\uparrow$ & $49.6^{\pm 0.4}$ & $54.4^{\pm 1.0}$ & $53.8^{\pm 1.0}$ & $55.0^{\pm 1.6}$ & $54.4^{\pm 1.7}$ \\
    & $FID_{\text{tr}}\downarrow$ & $22.54^{\pm 0.27}$ & $27.75^{\pm 1.05}$ & $27.98^{\pm 0.54}$ & $28.10^{\pm 0.52}$ & $28.27^{\pm 0.42}$\\
    & $FID_{\text{te}}\downarrow$ & $22.39^{\pm 0.36}$ & $27.97^{\pm 0.99}$ & $28.06^{\pm 0.60}$ & $28.32^{\pm 0.65}$ & $28.55^{\pm 0.51}$ \\
    & $Div_{\text{w}}\uparrow$ & $1.35^{\pm 0.00}$ & $1.32^{\pm 0.02}$ & $1.31^{\pm 0.01}$ & $1.29^{\pm 0.01}$ & $1.30^{\pm 0.01}$\\
    & $Div\uparrow$ & $1.74^{\pm 0.00}$ & $1.71^{\pm 0.02}$ & $1.69^{\pm 0.01}$ & $1.67^{\pm 0.01}$ & $1.68^{\pm 0.02}$\\
    \bottomrule
    \end{tabular}
    }
    \vspace{-0.3 cm}
    \caption{\textbf{Results on prediction with action label sequences.} Our model achieves stable performance at each prediction step.}\label{tab:actseq}
    \vspace{-0.6 cm}
\end{table}
\textbf{Trajectory smoothness.} We also compare the trajectories produced by different models in Fig.~\ref{fig:seq_traj}~(a). The motions predicted by Action2Motion~\cite{guo2020action2motion} suffer from heavy jitter, especially during the transition between the historical motion and the predicted one (as highlighted by the red circle). The reason is that Action2Motion employs a frame-wise random code, thus making the input to the decoder vary significantly across the frames. Note that this jitter makes our variance-based stopping criterion inapplicable to Action2Motion. We therefore tested different stopping strategies, detailed in the supplementary material, and report the one that gave the best results. When comparing our two models, we found the RNN-based one to produce smoother future motions than the Transformer-based one.
\vspace{-0.1cm}
\subsection{Ablation Study}
\vspace{-0.1cm}
\begin{figure*}[!ht]
\vspace{-0.3cm}
    \centering
    \includegraphics[width=0.8\linewidth]{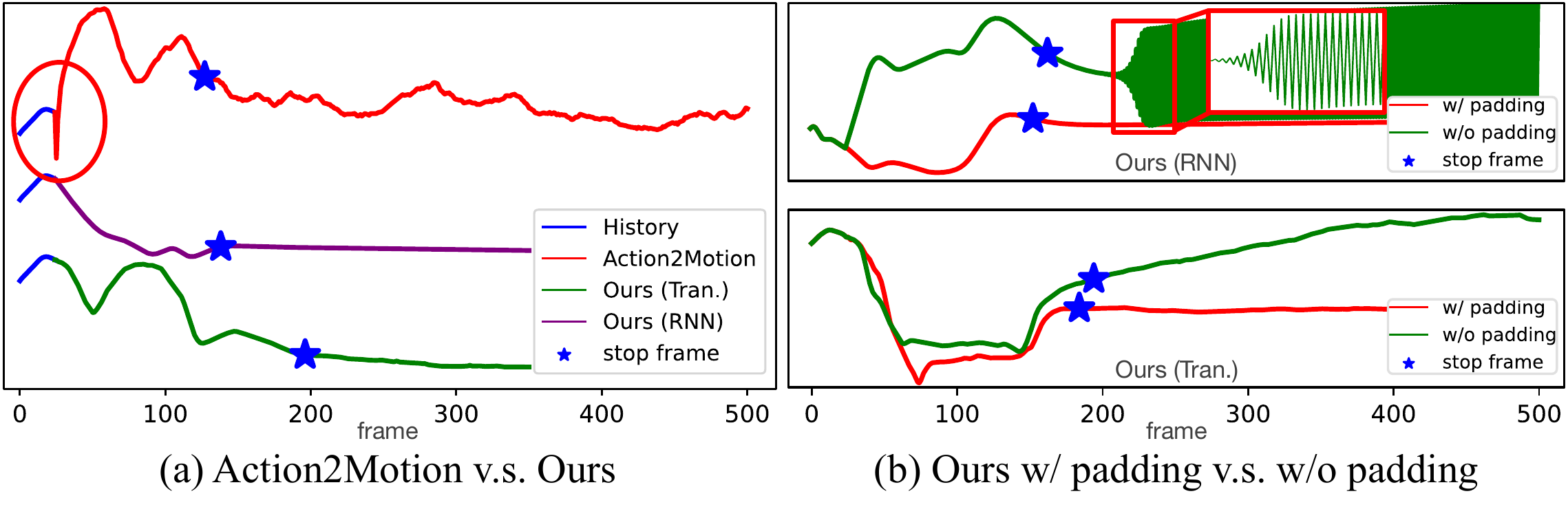}
    \vspace{-0.5cm}
    \caption{\textbf{Motion trajectories.} (a) The trajectories from our model are smoother than those of Action2Motion~\cite{guo2020action2motion}. (b) The trajectories from our model without ``padding'' either are unstable (top), or do not converge to static poses (bottom).}
    \label{fig:seq_traj}
    \vspace{-0.5cm}
\end{figure*}
\begin{table}[!ht]
    \centering
    \resizebox{\linewidth}{!}{
    \begin{tabular}{ccc|ccccc}
    \toprule
    &\multicolumn{2}{c|}{Method} & Acc$\uparrow$ & $FID_{\text{tr}}\downarrow$ & $FID_{\text{te}}\downarrow$ & $Div_{\text{w}}\uparrow$  & $Div\uparrow$\\
    \midrule
    \multirow{6}{*}{\rotatebox{90}{GRAB\;\;\;\;}}& \multirow{3}{*}{\rotatebox{90}{RNN\;\;\;}} 
    & w/o padding & $88.4^{\pm 0.6}$ & $\textbf{32.74}^{\pm 0.95}$ & $45.61^{\pm 1.62}$ & $\textbf{1.14}^{\pm 0.01}$ & $1.35^{\pm 0.01}$ \\
    & & w/o weakly-sup & $74.2^{\pm 0.8}$ & $93.84^{\pm 1.29}$ & $\textbf{11.42}^{\pm 1.26}$ & $0.20^{\pm 0.00}$ & $0.29^{\pm 0.00}$ \\\cmidrule{3-8}
    & & w/ both & $\textbf{92.6}^{\pm 0.6}$ & $44.59^{\pm 1.39}$ & $38.03^{\pm 1.49}$ & $1.10^{\pm 0.01}$ & $\textbf{1.37}^{\pm 0.01}$\\\cmidrule{2-8}
    & \multirow{3}{*}{\rotatebox{90}{Tran.\;\;\;}} 
    & w/o padding & $80.4^{\pm 0.5}$ & $\textbf{46.38}^{\pm 1.45}$  & $44.63^{\pm 1.19}$ & $\textbf{1.23}^{\pm 0.01}$ & $\textbf{1.12}^{\pm 0.00}$ \\
    & & w/o weakly-sup & $48.7^{\pm 0.8}$ & $184.86^{\pm 3.48}$ & $\textbf{23.03}^{\pm 1.29}$ & $0.01^{\pm 0.00}$ & $0.01^{\pm 0.00}$ \\\cmidrule{3-8}
    & & w/ both & $\textbf{85.5}^{\pm 1.2}$ & $48.58^{3.05}$ & $25.72^{\pm 2.16}$ & $1.05^{\pm 0.01}$ & $1.08^{\pm 0.01}$\\
    \bottomrule
    \multirow{6}{*}{\rotatebox{90}{NTU\;\;\;\;}}& \multirow{3}{*}{\rotatebox{90}{RNN\;\;\;}} 
    & w/o padding & $70.1^{\pm 0.2}$ & $119.25^{\pm 0.95}$ & $215.69^{\pm 2.43}$ & $\textbf{1.34}^{\pm 0.00}$ & $1.82^{\pm 0.01}$ \\
    & & w/o weakly-sup & $73.6^{\pm 0.3}$ & $107.88^{\pm 2.24}$ & $114.00^{\pm 0.71}$ & $0.53^{\pm 0.00}$ & $0.89^{\pm 0.01}$ \\\cmidrule{3-8}
    & & w/ both & $\textbf{76.0}^{\pm 0.2}$ & $\textbf{72.18}^{\pm 0.93}$  & $\textbf{111.01}^{\pm 1.28}$ & $1.25^{\pm 0.00}$ & $\textbf{2.20}^{\pm 0.00}$\\\cmidrule{2-8}
    & \multirow{3}{*}{\rotatebox{90}{Tran.\;\;\;}} & w/o padding & $69.1^{\pm 0.1}$ & $101.22^{\pm 1.65}$ & $118.44^{\pm 2.07}$ & $1.21^{\pm 0.00}$ & $1.62^{\pm 0.00}$ \\
    & & w/o weakly-sup & $64.7^{\pm 0.2}$ & $216.56^{\pm 2.96}$ & $264.92^{\pm 6.27}$ & $0.02^{\pm 0.00}$ & $0.02^{\pm 0.00}$ \\\cmidrule{3-8}
    & & w/ both & $\textbf{71.3}^{\pm 0.2}$ & $\textbf{83.14}^{\pm 1.74}$  & $\textbf{114.62}^{\pm 0.93}$ & $\textbf{1.25}^{\pm 0.00}$ & $\textbf{2.19}^{\pm 0.01}$\\
    \bottomrule
    \multirow{6}{*}{\rotatebox{90}{BABEL\;}}& \multirow{3}{*}{\rotatebox{90}{RNN\;\;\;}} 
    & w/o padding & $46.3^{\pm 0.2}$ & $39.08^{\pm 0.21}$ & $37.33^{\pm 0.29}$ & $\textbf{1.54}^{\pm 0.00}$ & $\textbf{1.78}^{\pm 0.00}$ \\
    & & w/o weakly-sup & $15.6^{\pm 0.1}$ & $\textbf{17.67}^{\pm 0.41}$ & $\textbf{15.57}^{\pm 0.41}$ & $0.05^{\pm 0.00}$ & $0.09^{\pm 0.00}$ \\\cmidrule{3-8}
    & & w/ both & $\textbf{49.6}^{\pm 0.4}$ & $22.54^{\pm 0.27}$ & $22.39^{\pm 0.36}$ & $1.35^{\pm 0.00}$ & $1.74^{\pm 0.00}$ \\\cmidrule{2-8}
    & \multirow{3}{*}{\rotatebox{90}{Tran.\;\;\;}} 
    & w/o padding & $37.8^{\pm 0.3}$ & $28.70^{\pm 0.31}$ & $27.64^{\pm 0.45}$ & $1.38^{\pm 0.00}$ & $1.61^{\pm 0.01}$ \\
    & & w/o weakly-sup & $12.3^{\pm 0.2}$ & $20.76^{\pm 0.26}$ & $\textbf{17.62}^{\pm 0.46}$ & $0.01^{\pm 0.00}$ & $0.01^{\pm 0.00}$ \\\cmidrule{3-8}
    & & w/ both & $\textbf{39.5}^{\pm 0.3}$ & $\textbf{20.02}^{\pm 0.24}$ & $19.41^{\pm 0.35}$ & $\textbf{1.39}^{\pm 0.00}$ & $\textbf{1.82}^{\pm 0.01}$ \\
    \bottomrule
    \end{tabular}
    }
    \vspace{-0.3 cm}
    \caption{\textbf{Ablation studies} on generating additional static frames for variable length prediction (padding) and weakly-supervised action transition learning (weakly-sup). Note that, without ``weakly-sup'', the Transformer-based models suffer from mode collapse, leading to very low motion diversity.}
    \label{tab:ablation}
    \vspace{-0.3 cm}
\end{table}
\begin{table}[!ht]
    \centering
    \resizebox{\linewidth}{!}{
    \begin{tabular}{cc|ccccc}
    \toprule
    \multicolumn{2}{c|}{Method} & Acc$\uparrow$ & $FID_{\text{tr}}\downarrow$ & $FID_{\text{te}}\downarrow$ & $Div_{\text{w}}\uparrow$  & $Div\uparrow$\\
    \midrule
    \multirow{4}{*}{\rotatebox{90}{RNN\;\;\;}} 
    & w/o both & $15.6^{\pm 0.1}$ & $\textbf{17.67}^{\pm 0.41}$ & $\textbf{15.57}^{\pm 0.41}$ & $0.05^{\pm 0.00}$ & $0.09^{\pm 0.00}$ \\
    & w/ gt-transi & $16.4^{\pm 0.4}$ & $21.28^{\pm 0.31}$ & $18.83^{\pm 0.34}$ & $0.07^{\pm 0.00}$ & $0.11^{\pm 0.00}$ \\
    & w/ weakly-sup & $\textbf{49.6}^{\pm 0.4}$ & $22.54^{\pm 0.27}$ & $22.39^{\pm 0.36}$ & $\textbf{1.35}^{\pm 0.00}$ & $\textbf{1.74}^{\pm 0.00}$ \\
    & w/ both & $48.4^{\pm 0.5}$ & $22.70^{\pm 0.23}$ & $22.47^{\pm 0.33}$ & $1.31^{\pm 0.00}$ & $1.71^{\pm 0.01}$ \\\cmidrule{1-7}
    \multirow{4}{*}{\rotatebox{90}{Tran.\;\;\;}} 
    & w/o both & $12.3^{\pm 0.2}$ & $20.76^{\pm 0.26}$ & $17.62^{\pm 0.46}$ & $0.01^{\pm 0.00}$ & $0.01^{\pm 0.00}$ \\
    & w/ gt-transi & $13.0^{\pm 0.3}$ & $20.40^{\pm 0.37}$ & $\textbf{17.56}^{\pm 0.48}$ & $0.01^{\pm 0.00}$ & $0.01^{\pm 0.00}$ \\
    & w/ weakly-sup & $\textbf{39.5}^{\pm 0.3}$ & $\textbf{20.02}^{\pm 0.24}$ & $19.41^{\pm 0.35}$ & $1.39^{\pm 0.00}$ & $1.82^{\pm 0.01}$ \\
    & w/ both & $38.5^{\pm 0.3}$ & $20.79^{\pm 0.27}$ & $20.12^{\pm 0.39}$ & $\textbf{1.41}^{\pm 0.00}$ & $\textbf{1.86}^{\pm 0.01}$ \\
    \bottomrule
    \end{tabular}
    }
    \vspace{-0.3 cm}
    \caption{\textbf{Ablation studies} on training with ground truth transition v.s. our weakly-supervised action transition learning.}
    \label{tab:abla-gttransi}
    \vspace{-0.5 cm}
\end{table}
To provide a deeper understanding of our model, we evaluate the influence of its two main components, i.e., encouraging the model to predict static poses at the end of the sequence, which we refer to as ``padding'', and weakly-supervised action transition learning (``weakly-sup''). The results are shown in Table~\ref{tab:ablation}. In general, the model with both components achieves the best performance across all datasets for both temporal encoding structures. Although the numerical results of the models without padding are close to those with padding, we observed the trajectories generated by such models to occasionally either be unstable (RNN-based model) or not converge to a static pose (Transformer-based model), as shown in Fig.~\ref{fig:seq_traj}~(b). Without our weakly-supervised transition learning, the models often fail to produce diverse future motions and the Transformer-based model suffers from mode collapse.

Finally, we compare the performance of using the ground-truth transitions (``gt-transi'') to that of our weakly-supervised strategy (``weakly-sup'') on BABEL in Table~\ref{tab:abla-gttransi}. Since the limited ground-truth transitions in BABEL do not cover all possible cases, using them as supervision is ineffective. Specifically, as shown in Table~\ref{tab:dataset}, there are only around 2500 ground-truth transition sequences, depicting 170 types of transitions. By contrast, our weakly-supervised strategy leverages almost 100,000 pseudo transitions covering all 380 possible types. This further evidences the importance of our weakly-supervised action transition learning strategy.
\vspace{-0.2cm}
\section{Conclusion}\label{sec:conclu}
\vspace{-0.2cm}
In this paper, we have introduced the task of \emph{action-driven stochastic human motion prediction}, which aims to predict future trajectories of a given action category. Since it is unrealistic to expect a human motion dataset to include all possible action transitions, we have introduced a weakly-supervised training procedure to learn those transitions from a dataset with only a single action label per sequence. Furthermore, we have introduced a variance-based strategy to produce motions of variable length. Our current model can only generate motions of actions observed in the training set, thus not allowing us to explore novel actions at test time. We will seek to address this in our future work. 

\noindent{\textbf{Limitations \& Negative Societal Impacts}}

One limitation of our work arises from the fact that our model does not predict global translations. The human movements include local body motions and global translations. However, without scene context, we cannot ensure a valid global translation. For example, a ``sit'' motion needs to result in sitting on a chair (or something) of the scene.

A potential risk of applying our method to real scenario is that, without considering the scene context, the predicted human motions can lead to unsafe situations, such as collision. We recommend to validate the outputs of our model w.r.t. the environment before applying them to robots/agents.

\noindent{\textbf{Acknowledgements}}

This research was supported in part by the Australia Research Council DECRA Fellowship (DE180100628) and ARC Discovery Grant (DP200102274). The authors would like to thank NVIDIA for the donated GPU (Titan V).

{\small
\bibliographystyle{ieee_fullname}
\bibliography{action-driven-motion-pred}
}

\pagebreak
\setcounter{equation}{0}
\setcounter{figure}{0}
\setcounter{table}{0}
\setcounter{page}{1}
\twocolumn[{
	\begin{@twocolumnfalse}
		\begin{center}
			\textbf{\Large Weakly-supervised Action Transition Learning for \\Stochastic Human Motion Prediction\\--- Supplementary Material ---\\}
			
			\author{\; \\
				Wei Mao$^1$, \;\;Miaomiao Liu$^{1}$,\;\; Mathieu Salzmann$^{2,3}$\;\; \\
				$^1$Australian National University; $^2$CVLab, EPFL; $^3$ClearSpace, Switzerland\\
				{\tt\small \{wei.mao, miaomiao.liu\}@anu.edu.au,}\;\;{\tt\small mathieu.salzmann@epfl.ch}
			}
		\end{center}
	\end{@twocolumnfalse}
}]

\section{Weakly-supervised Transition Learning}
We illustrate our weakly-supervised transition learning in Fig.~\ref{fig:weakly-sup} below. Given past $N$ frames $\textbf{X}_{1:N}$, our model will predict future $T+T_0+P$ frames. The first $T_0$ frames are the transition between the past and future motion where a weak supervision signal ($\mathcal{L}_{\text{smooth}}$) is applied during training. The last $P$ frames are supervised by the repeats of last pose in future motion included in the reconstruction loss ($\mathcal{L}_{\text{rec}}$). This encourages the model to predict static poses after predicting the future motion.
\begin{figure}[ht]
	\centering
	\includegraphics[width=\linewidth]{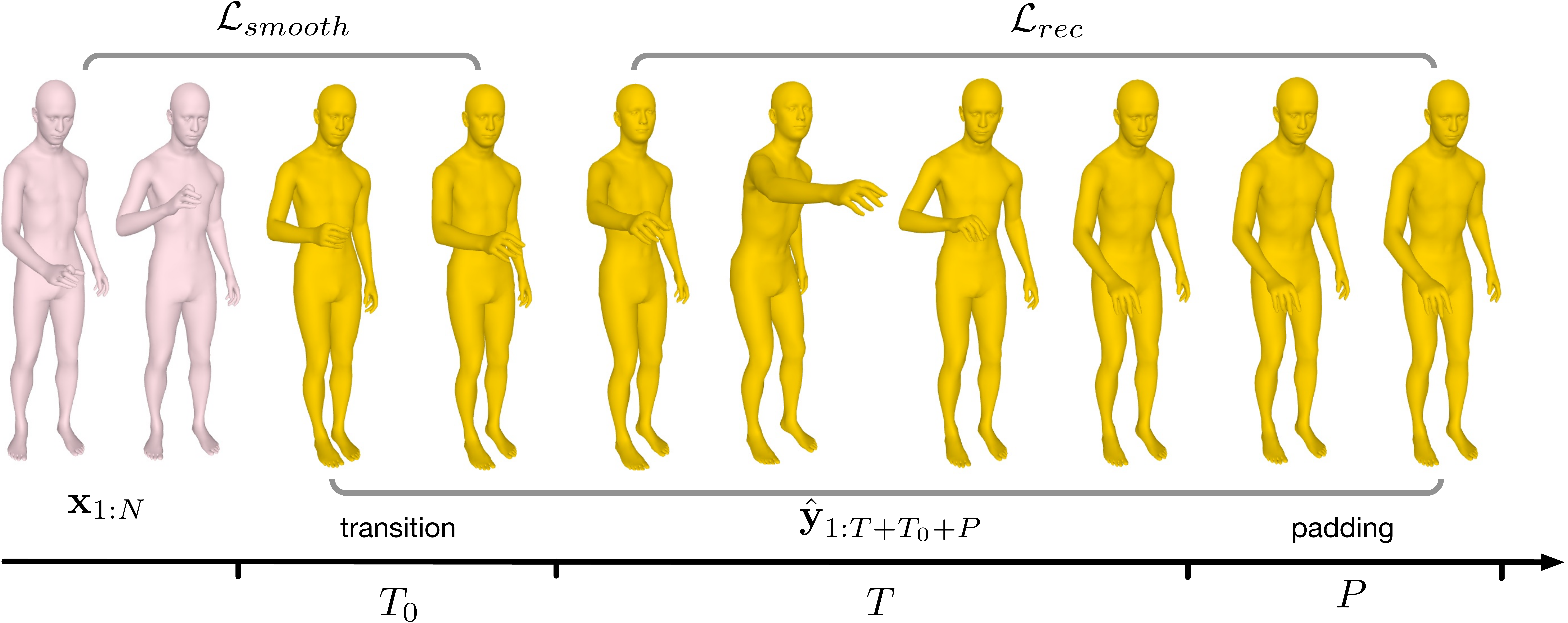}
	\caption{An illustration of our weakly supervision transition learning.}
	\label{fig:weakly-sup}
\end{figure}

\section{Results on HumanAct12}
\noindent\textbf{HumanAct12} (HAct)~\cite{guo2020action2motion}  is a subset of the PHSPD dataset~\cite{zou20203d}. It consists of 12 subjects performing 12 actions. We use 2 subjects (P11, P12) for testing and the remaining 10 subjects (P1-P10) for training. The minimum and maximum length of the motions are 35 and 290 frames, respectively. We remove the motions that are too short, i.e., less than 35 frames, leading to 727 training sequences and 197 testing ones. Our model is trained to observe 10 frames. 

The results on HAct~\cite{guo2020action2motion} are shown in Tables~\ref{tab:hact},~\ref{tab:hact-ade} and~\ref{tab:hact-actseq}.
\begin{table}[!ht]
	\vspace{-0.2cm}
	\centering
	\resizebox{\linewidth}{!}{
		\begin{tabular}{cc|ccccc}
			\toprule
			&Method & Acc$\uparrow$ & $FID_{\text{tr}}\downarrow$ & $FID_{\text{te}}\downarrow$ & $Div_{\text{w}}\uparrow$  & $Div\uparrow$\\
			\midrule
			
			\multirow{5}{*}{\rotatebox{90}{HAct}}
			&Act2Mot~\cite{guo2020action2motion} & $24.5^{\pm 0.1}$ & $245.35^{\pm 7.13}$ & $298.06^{\pm 10.80}$ & $0.31^{\pm 0.00}$ & $0.60^{\pm 0.01}$ \\
			&DLow~\cite{yuan2020dlow} & $22.7^{\pm 0.2}$ & $254.72^{\pm 8.48}$ & $143.71^{\pm 3.07}$ & $0.35^{\pm 0.00}$ & $0.53^{\pm 0.00}$ \\
			&ACTOR~\cite{petrovich21actor} & $44.4^{\pm 0.2}$ & $248.81^{\pm 3.77}$ & $381.56^{\pm 6.66}$ & $0.84^{\pm 0.00}$ & $0.95^{\pm 0.00}$ \\\cmidrule{2-7}
			&Ours (RNN) & $\textbf{59.0}^{\pm 0.1}$ & $\textbf{129.95}^{\pm 0.39}$ & $164.38^{\pm 2.27}$ & $\textbf{0.74}^{\pm 0.00}$ & $\textbf{0.96}^{\pm 0.00}$ \\
			&Ours (Tran.) & $56.8^{\pm 0.2}$ & $141.85^{\pm 2.51}$ & $\textbf{139.82}^{\pm 1.80}$ & $0.67^{\pm 0.00}$ & $0.88^{\pm 0.00}$ \\
			\bottomrule
		\end{tabular}
	}
	\caption{\textbf{Quantitative results on HumanAct12~\cite{guo2020action2motion}.} We adapt Action2Motion~\cite{guo2020action2motion}, ACTOR~\cite{petrovich21actor} and DLow~\cite{yuan2020dlow} to our task.
	}\label{tab:hact}
\end{table}

\begin{table}[!ht]
	\centering
	\resizebox{0.5\linewidth}{!}{
		\begin{tabular}{c|cc}
			\toprule
			& \multicolumn{2}{c}{HAct}  \\
			Method & $ADE_{\text{w}}\downarrow$  & $ADE\downarrow$ \\
			\midrule
			Act2Mot~\cite{guo2020action2motion} & $1.09^{\pm 0.00}$ & $1.38^{\pm 0.01}$  \\
			DLow~\cite{yuan2020dlow} & $1.15^{\pm 0.01}$ & $1.39^{\pm 0.01}$ \\
			ACTOR~\cite{petrovich21actor} & $1.21^{\pm 0.02}$ & $1.54^{\pm 0.01}$ \\\midrule
			Ours (RNN) & $1.06^{\pm 0.01}$ & $\textbf{1.23}^{\pm 0.01}$ \\
			Ours (Tran.) & $\textbf{1.02}^{\pm 0.01}$ & $1.26^{\pm 0.01}$ \\
			\bottomrule
		\end{tabular}
	}
	\caption{\textbf{Results of prediction accuracy with the ground truth action label.}}\label{tab:hact-ade}
\end{table}

\begin{table}[!ht]
	\vspace{-0.3 cm}
	\centering
	\resizebox{\linewidth}{!}{
		\begin{tabular}{cc|ccccc}
			\toprule
			& & \multicolumn{5}{c}{Prediction Step} \\
			& Metrics & $1_{st}$  & $2_{nd}$ & $3_{rd}$  & $4_{th}$ & $5_{th}$ \\
			\midrule
			\multirow{5}{*}{\rotatebox{90}{HAct\;\;}} 
			& Acc$\uparrow$ & $59.0^{\pm 0.1}$ & $60.9^{\pm 1.0}$ & $60.9^{\pm 0.7}$ & $60.8^{\pm 0.8}$ & $60.3^{\pm 0.5}$ \\
			& $FID_{\text{tr}}\downarrow$ & $129.95^{\pm 0.39}$ & $148.49^{\pm 9.44}$ & $159.06^{\pm 10.27}$ & $158.88^{\pm 8.93}$ & $166.06^{\pm 9.41}$ \\
			& $FID_{\text{te}}\downarrow$ & $164.38^{\pm 2.27}$ & $240.23^{\pm 7.57}$ & $259.46^{\pm 7.52}$ & $258.19^{\pm 13.83}$ & $262.20^{\pm 19.21}$ \\
			& $Div_{\text{w}}\uparrow$ & $0.74^{\pm 0.00}$ & $0.81^{\pm 0.01}$ & $0.80^{\pm 0.01}$ & $0.81^{\pm 0.01}$ & $0.81^{\pm 0.00}$ \\
			& $Div\uparrow$ & $0.96^{\pm 0.00}$ & $1.05^{\pm 0.01}$ & $1.04^{\pm 0.02}$ & $1.06^{\pm 0.01}$ & $1.06^{\pm 0.01}$ \\ 
			\bottomrule
		\end{tabular}
	}
	\caption{\textbf{Results on prediction with action label sequences.}}\label{tab:hact-actseq}
\end{table}

\section{Implementation Details}
In this section, we describe the implementation details for each dataset and model. In Tables~\ref{tab:imple-rnn} and~\ref{tab:imple-trans}, we show the detailed network design and hyperparameters used in our experiments for the RNN-based and Transformer-based models, respectively.

\begin{table}[!ht]
	\centering
	\resizebox{0.7\linewidth}{!}{
		\begin{tabular}{c|cc|cccc}
			\toprule
			\multirow{2}{*}{Dataset}&    \multicolumn{2}{c|}{Feature Size} & \multirow{2}{*}{$\lambda_{\text{rec}}$} & \multirow{2}{*}{$\lambda_{\text{smooth}}$} & \multirow{2}{*}{LR} & \multirow{2}{*}{$P$} \\
			& GRU & MLP &  &  &  & \\
			\midrule
			GRAB & 128 & [300, 200, 128] & 100.0 & 100.0 & 0.002 & 50\\
			NTU & 128 & [300, 200, 128] & 100.0 & 20.0 & 0.002 & 50\\
			BABEL & 256 & [512, 256, 256] & 50.0 & 10.0 & 0.001 & 50 \\
			HAct & 128 & [300, 200, 128] & 100.0 & 100.0 & 0.002 & 50\\\bottomrule
		\end{tabular}
	}
	\caption{\textbf{Implementation details of our RNN-based model.} We show the network design, loss weights ($\lambda_{\text{rec}}$ and $\lambda_{\text{smooth}}$), learning rate (LR) and number of padding frames ($P$). Note that the posterior and prior modules share the same network architecture.}
	\label{tab:imple-rnn}
\end{table}

\begin{table}[!ht]
	\centering
	\resizebox{\linewidth}{!}{
		\begin{tabular}{c|cccc|cccc}
			\toprule
			\multirow{2}{*}{Dataset}& \multicolumn{4}{c|}{Transformer} & \multirow{2}{*}{$\lambda_{\text{rec}}$} & \multirow{2}{*}{$\lambda_{\text{smooth}}$} & \multirow{2}{*}{LR} & \multirow{2}{*}{$P$} \\
			& d\_model & nhead  & dim\_feedforward & num\_layers &  &  &  & \\
			\midrule
			GRAB & 128 & 4 & 256 & 6 & 1000.0 & 100.0 & 0.0005 & 20\\
			NTU & 128 & 4 & 512 & 8 & 100.0 & 20.0 & 0.0005 & 20\\
			BABEL & 128 & 4 & 256 & 6 & 100.0 & 10.0 & 0.0001 & 20 \\
			HAct & 128 & 4 & 256 & 6 & 1000.0 & 100.0 & 0.0005 & 50\\\bottomrule
		\end{tabular}
	}
	\caption{\textbf{Implementation details of our Transformer-based model.}}
	\label{tab:imple-trans}
\end{table}

\textbf{Training details.} We mainly train our RNN-based models on an NVIDIA TITAN-V GPU. Training for 500 epochs takes 2-12 hours. For Transformer-based models, since they consume much more GPU memory, we train them on an NVIDIA RTX3090 GPU, which takes 5-60 hours.

\begin{table}[!ht]
	\centering
	\resizebox{\linewidth}{!}{
		\begin{tabular}{ccc|ccccc}
			\toprule
			&\multicolumn{2}{c|}{Method} & Acc$\uparrow$ & $FID_{\text{tr}}\downarrow$ & $FID_{\text{te}}\downarrow$ & $Div_{\text{w}}\uparrow$  & $Div\uparrow$\\
			\midrule
			\multirow{4}{*}{\rotatebox{90}{GRAB\;\;\;\;\;}}& \multirow{2}{*}{\rotatebox{90}{RNN\;}} 
			& MLP & $\textbf{93.5}^{\pm 0.6}$ & $\textbf{33.18}^{\pm 1.27}$ & $43.86^{\pm 1.08}$ & $\textbf{1.16}^{\pm 0.01}$ & $\textbf{1.39}^{\pm 0.01}$ \\\cmidrule{3-8}
			& & \CC{10}linear func. & \CC{10}$92.6^{\pm 0.6}$ & \CC{10}$44.59^{\pm 1.39}$ & \CC{10}$\textbf{38.03}^{\pm 1.49}$ & \CC{10}$1.10^{\pm 0.01}$ & \CC{10}$1.37^{\pm 0.01}$ \\\cmidrule{2-8}
			& \multirow{2}{*}{\rotatebox{90}{Tran.\;}} 
			& MLP & $65.1^{\pm 1.2}$ & $167.14^{\pm 3.23}$ & $55.41^{\pm 2.08}$ & $0.02^{\pm 0.00}$ & $0.00^{\pm 0.00}$ \\\cmidrule{3-8}
			& & \CC{10}linear func. & \CC{10}$\textbf{85.5}^{\pm 1.2}$ & \CC{10}$\textbf{48.58}^{3.05}$ & \CC{10}$\textbf{25.72}^{\pm 2.16}$ & \CC{10}$\textbf{1.05}^{\pm 0.01}$ & \CC{10}$\textbf{1.08}^{\pm 0.01}$ \\
			\bottomrule
			\multirow{4}{*}{\rotatebox{90}{NTU\;\;}}& \multirow{2}{*}{\rotatebox{90}{RNN\;}} 
			& MLP & $\textbf{76.3}^{\pm 0.2}$ & $79.55^{\pm 1.32}$ & $113.61^{\pm 0.89}$ & $\textbf{1.36}^{\pm 0.00}$ & $\textbf{2.20}^{\pm 0.00}$ \\\cmidrule{3-8}
			& & \CC{10}linear func. & \CC{10}$76.0^{\pm 0.2}$ & \CC{10}$\textbf{72.18}^{\pm 0.93}$  & \CC{10}$\textbf{111.01}^{\pm 1.28}$ & \CC{10}$1.25^{\pm 0.00}$ & \CC{10}$\textbf{2.20}^{\pm 0.00}$ \\\cmidrule{2-8}
			& \multirow{2}{*}{\rotatebox{90}{Tran.\;}} 
			& MLP & $61.2^{\pm 0.1}$ & $316.15^{\pm 6.04}$ & $237.20^{\pm 3.59}$ & $0.00^{\pm 0.00}$ & $0.01^{\pm 0.00}$ \\\cmidrule{3-8}
			& & \CC{10}linear func. & \CC{10}$\textbf{71.3}^{\pm 0.2}$ & \CC{10}$\textbf{83.14}^{\pm 1.74}$  & \CC{10}$\textbf{114.62}^{\pm 0.93}$ & \CC{10}$\textbf{1.25}^{\pm 0.00}$ & \CC{10}$\textbf{2.19}^{\pm 0.01}$ \\
			\bottomrule
			\multirow{4}{*}{\rotatebox{90}{BABEL\;\;}}& \multirow{2}{*}{\rotatebox{90}{RNN\;}}
			& MLP & $\textbf{54.4}^{\pm 0.4}$ & $\textbf{20.86}^{\pm 0.29}$ & $\textbf{21.46}^{\pm 0.35}$ & $\textbf{1.55}^{\pm 0.00}$ & $\textbf{1.78}^{\pm 0.00}$ \\\cmidrule{3-8}
			& & \CC{10}linear func. & \CC{10}$49.6^{\pm 0.4}$ & \CC{10}$22.54^{\pm 0.27}$ & \CC{10}$22.39^{\pm 0.36}$ & \CC{10}$1.35^{\pm 0.00}$ & \CC{10}$1.74^{\pm 0.00}$\\\cmidrule{2-8}
			& \multirow{2}{*}{\rotatebox{90}{Tran.\;}} 
			& MLP & $37.9^{\pm 0.5}$ & $44.05^{\pm 0.68}$ & $41.92^{\pm 0.71}$ & $0.00^{\pm 0.00}$ & $0.00^{\pm 0.00}$ \\\cmidrule{3-8}
			& & \CC{10}linear func. & \CC{10}$\textbf{39.5}^{\pm 0.3}$ & \CC{10}$\textbf{20.02}^{\pm 0.24}$ & \CC{10}$\textbf{19.41}^{\pm 0.35}$ & \CC{10}$\textbf{1.39}^{\pm 0.00}$ & \CC{10}$\textbf{1.82}^{\pm 0.01}$
			\\\bottomrule
			\multirow{4}{*}{\rotatebox{90}{HAct\;\;}}& \multirow{2}{*}{\rotatebox{90}{RNN\;}} 
			& MLP & $\textbf{61.0}^{\pm 0.0}$ & $140.34^{\pm 1.15}$ & $\textbf{142.59}^{\pm 2.36}$ & $\textbf{0.88}^{\pm 0.00}$ & $\textbf{1.06}^{\pm 0.00}$ \\\cmidrule{3-8}
			& & \CC{10}linear func. & \CC{10}$59.0^{\pm 0.1}$ & \CC{10}$\textbf{129.95}^{\pm 0.39}$ & \CC{10}$164.38^{\pm 2.27}$ & \CC{10}$0.74^{\pm 0.00}$ & \CC{10}$0.96^{\pm 0.00}$ \\\cmidrule{2-8}
			& \multirow{2}{*}{\rotatebox{90}{Tran.\;}} 
			& MLP & $47.5^{\pm 0.3}$ & $217.63^{\pm 3.33}$ & $206.94^{\pm 2.60}$ & $0.01^{\pm 0.00}$ & $0.01^{\pm 0.00}$ \\\cmidrule{3-8}
			& & \CC{10}linear func. & \CC{10}$\textbf{56.8}^{\pm 0.2}$ & $\CC{10}\textbf{141.85}^{\pm 2.51}$ & \CC{10}$\textbf{139.82}^{\pm 1.80}$ & $\CC{10}\textbf{0.67}^{\pm 0.00}$ & \CC{10}$\textbf{0.88}^{\pm 0.00}$ \\
			\bottomrule
		\end{tabular}
	}
	\caption{\textbf{Ablation study} on the function to produce $T_0$. The model we choose is highlighted with a blue background.}
	\label{tab:func_t0}
\end{table}

\section{Functions for $T_0$}
As described in the main paper, we formulate the number of transition frames $T_0$ between pose sequences from two different motions as a function
\begin{align}
	T_0 = f(\mathbf{x}_N,\mathbf{y}'_{1})\;,   
\end{align}
where $\mathbf{x}_N$ and $\mathbf{y}'_{1}$ are the last pose of the historical sequences~($\mathbf{X}$) and the first pose of future sequences~($\mathbf{Y}'$), respectively.

We then define this as a simple linear function
\begin{align}
	T_0 = \lfloor {k\|\mathbf{x}_N-\mathbf{y}'_{1}\|_2}\rfloor\;,
\end{align}
where $k\geq 0$ is calculated from the training data. More specifically, given pose sequences $\mathbf{X}=[\mathbf{x}_1,\mathbf{x}_2,\cdots]$ from the training set~($\mathcal{D}$), $k$ is computed as
\begin{align}
	k = \underset{\mathbf{X}\in\mathcal{D}, i\neq j}{\mathbb{E}}\left[\frac{|i-j|}{\|\mathbf{x}_{i}-\mathbf{x}_{j}\|_2}\right]\;.
\end{align}

To demonstrate the effectiveness of such a simple linear function, we learn a more complicated one, modeled by a multilayer perception (MLP), from training data. As shown in Table~\ref{tab:func_t0}, for RNN-based models both functions perform on par with each other. By contrast, for Transformer-based models, the linear function yields better performance than the MLP one.

\begin{figure}[!t]
	\centering
	\includegraphics[width=0.5\linewidth]{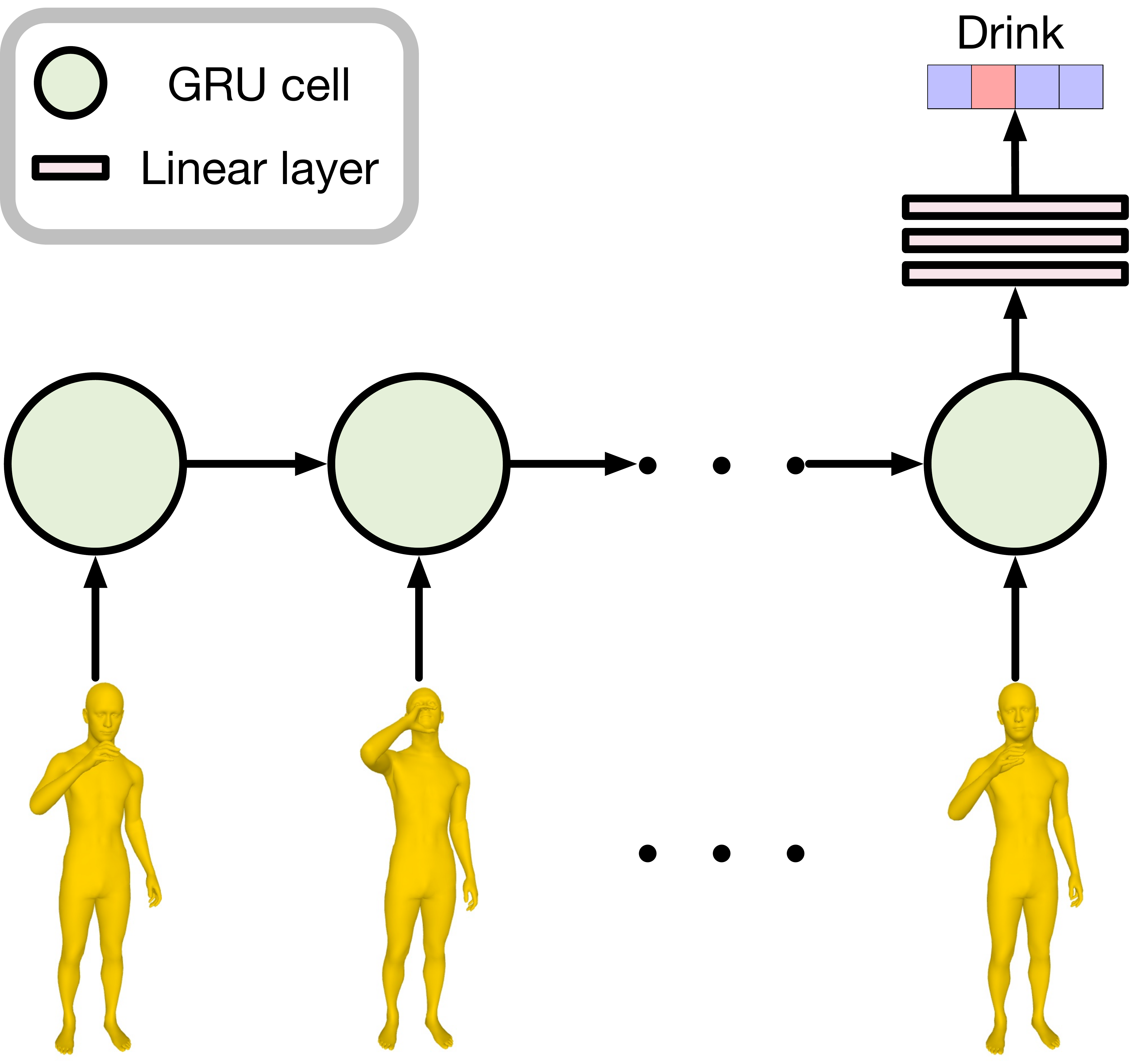} 
	\caption{{\bf Network structure} of the action recognition model. }
	\label{fig:act_class}
\end{figure}

\section{Action Recognition Model}

\begin{table}[!ht]
	\centering
	\resizebox{0.7\linewidth}{!}{
		\begin{tabular}{c|cccc}
			\toprule
			Dataset & GRAB & NTU & BABEL & HAct\\\midrule
			Acc & $85.5$ & $89.5$ & $60.8$ & $70.4$  \\\midrule
			$FID_{\text{tr}}$ & $116.46$ & $127.44$ & $9.62$ & $127.77$ \\\bottomrule
		\end{tabular}
	}
	\caption{\textbf{Results of action recognition model.} We report the action recognition accuracy (Acc) on the test set, the FID of test set to training data ($FID_{\text{tr}}$).}\label{tab:results_testset}
\end{table}
In Fig.~\ref{fig:act_class}, we provide the network structure of the action recognition model used for evaluation. It consists of a GRU layer to encode the temporal information and of a 3-layer MLP to produce the final classification results. Each layer of the MLP is followed by a ReLU~\cite{nair2010rectified} activation function, except for the last layer, which is followed by a sigmoid function. The model is then trained for 500 epochs using the ADAM~\cite{kingma2014adam} optimizer with an initial learning rate of 0.002.

To compute the Fréchet Inception Distance (FID)~\cite{heusel2017gans}, we take the features generated by the penultimate layer of the pretrained MLP. We then collect three sets of features computed from the training data, the testing data and generated motions, respectively. Each set of future motions is summarized as a multivariate Gaussian by calculating the mean and covariance matrix of these features. The FID then measures the distance between pairs of Gaussian distributions. We report the action recognition results, as well as the FID between the test set and the training data in Table~\ref{tab:results_testset}.

\begin{table}[!ht]
	\centering
	\resizebox{\linewidth}{!}{
		\begin{tabular}{ccc|ccccc}
			\toprule
			& & $P$ & Acc$\uparrow$ & $FID_{\text{tr}}\downarrow$ & $FID_{\text{te}}\downarrow$ & $Div_{\text{w}}\uparrow$  & $Div\uparrow$\\
			\midrule
			\multirow{6}{*}{\rotatebox{90}{GRAB\;\;\;\;}}& \multirow{3}{*}{\rotatebox{90}{RNN\;\;\;}} 
			& $10$ & $91.0^{\pm 0.4}$ & $54.61^{\pm 1.32}$ & $\textbf{29.86}^{\pm 0.69}$ & $\textbf{1.18}^{\pm 0.01}$ & $1.34^{\pm 0.01}$ \\
			& & $20$ & $90.6^{\pm 0.6}$ & $\textbf{37.79}^{\pm 1.05}$ & $30.80^{\pm 0.90}$ & $1.11^{\pm 0.01}$ & $1.34^{\pm 0.01}$ \\
			& & \CC{10}$50$ & \CC{10}$\textbf{92.6}^{\pm 0.6}$ & \CC{10}$44.59^{\pm 1.39}$ & \CC{10}$38.03^{\pm 1.49}$ & \CC{10}$1.10^{\pm 0.01}$ & \CC{10}$\textbf{1.37}^{\pm 0.01}$ \\\cmidrule{2-8}
			& \multirow{3}{*}{\rotatebox{90}{Tran.\;\;\;}} 
			& $10$ & $77.4^{\pm 1.0}$ & $75.65^{\pm 2.08}$ & $\textbf{17.12}^{\pm 0.93}$ & $\textbf{1.16}^{\pm 0.01}$ & $1.12^{\pm 0.01}$ \\
			& & \CC{10}$20$ & \CC{10}$\textbf{85.5}^{\pm 1.2}$ & \CC{10}$48.58^{\pm 3.05}$ & \CC{10}$25.72^{\pm 2.16}$ & \CC{10}$1.05^{\pm 0.01}$ & \CC{10}$1.08^{\pm 0.01}$ \\
			& & $50$ & $85.3^{\pm 0.6}$ & $\textbf{41.71}^{\pm 2.07}$ & $30.91^{\pm 1.30}$ & $1.12^{\pm 0.01}$ & $\textbf{1.22}^{\pm 0.01}$ \\
			\bottomrule
			\multirow{6}{*}{\rotatebox{90}{NTU\;\;\;\;}}& \multirow{3}{*}{\rotatebox{90}{RNN\;\;\;}} 
			& $10$ & $75.6^{\pm 0.1}$ & $76.53^{\pm 1.22}$ & $126.15^{\pm 1.28}$ & $\textbf{1.27}^{\pm 0.00}$ & $2.05^{\pm 0.01}$ \\
			& & $20$ & $75.9^{\pm 0.1}$ & $75.64^{\pm 1.04}$ & $120.30^{\pm 0.93}$ & $1.23^{\pm 0.00}$ & $2.12^{\pm 0.01}$ \\
			& & \CC{10}$50$ & \CC{10}$\textbf{76.0}^{\pm 0.2}$ & \CC{10}$\textbf{72.18}^{\pm 0.93}$ & \CC{10}$\textbf{111.01}^{\pm 1.28}$ & \CC{10}$1.25^{\pm 0.00}$ & \CC{10}$\textbf{2.20}^{\pm 0.01}$ \\\cmidrule{2-8}
			& \multirow{3}{*}{\rotatebox{90}{Tran.\;\;\;}} 
			& $10$ & $71.1^{\pm 0.1}$ & $88.53^{\pm 1.25}$ & $127.21^{\pm 1.48}$ & $\textbf{1.27}^{\pm 0.00}$ & $1.88^{\pm 0.01}$ \\
			& & \CC{10}$20$ & \CC{10}$\textbf{71.3}^{\pm 0.2}$ & \CC{10}$\textbf{83.14}^{\pm 1.74}$ & \CC{10}$\textbf{114.62}^{\pm 0.93}$ & \CC{10}$1.25^{\pm 0.00}$ & \CC{10}$\textbf{2.19}^{\pm 0.01}$ \\
			& & $50$ & $57.3^{\pm 0.1}$ & $478.00^{\pm 6.77}$ & $328.87^{\pm 3.98}$ & $0.00^{\pm 0.00}$ & $0.00^{\pm 0.00}$ \\
			\bottomrule
			
			\multirow{6}{*}{\rotatebox{90}{BABEL\;\;\;\;}}& \multirow{3}{*}{\rotatebox{90}{RNN\;\;\;}} 
			& $10$ & $48.8^{\pm 0.2}$ & $33.40^{\pm 0.26}$ & $32.55^{\pm 0.35}$ & $\textbf{1.48}^{\pm 0.00}$ & $\textbf{1.71}^{\pm 0.00}$\\
			& & $20$ & $\textbf{49.8}^{\pm 0.3}$ & $27.05^{\pm 0.25}$ & $26.56^{\pm 0.34}$ & $1.42^{\pm 0.00}$ & $\textbf{1.71}^{\pm 0.01}$\\
			& & \CC{10}$50$ & \CC{10}$49.6^{\pm 0.4}$ & \CC{10}$\textbf{22.54}^{\pm 0.27}$ & \CC{10}$\textbf{22.39}^{\pm 0.36}$ & \CC{10}$1.35^{\pm 0.00}$ & \CC{10}$1.74^{\pm 0.00}$\\\cmidrule{2-8}
			& \multirow{3}{*}{\rotatebox{90}{Tran.\;\;\;}} 
			& $10$ & $38.6^{\pm 0.3}$ & $22.78^{\pm 0.31}$ & $22.03^{\pm 0.42}$ & $\textbf{1.45}^{\pm 0.01}$ & $1.77^{\pm 0.01}$\\
			& & \CC{10}$20$ & \CC{10}$\textbf{39.5}^{\pm 0.3}$ & \CC{10}$20.02^{\pm 0.24}$ & \CC{10}$19.41^{\pm 0.35}$ & \CC{10}$1.39^{\pm 0.00}$ & \CC{10}$\textbf{1.82}^{\pm 0.01}$\\
			& & $50$ & $37.1^{\pm 0.3}$ & $\textbf{19.65}^{\pm 0.27}$ & $\textbf{18.93}^{\pm 0.37}$ & $1.20^{\pm 0.00}$ & $1.73^{\pm 0.01}$\\
			\bottomrule
			
			\multirow{6}{*}{\rotatebox{90}{HAct\;}}& \multirow{3}{*}{\rotatebox{90}{RNN\;\;\;}} 
			& $10$ & $50.3^{\pm 0.1}$ & $159.19^{\pm 2.58}$ & $164.58^{\pm 2.97}$ & $0.78^{\pm 0.00}$ & $\textbf{1.06}^{\pm 0.00}$ \\
			& & $20$ & $58.1^{\pm 0.2}$ & $\textbf{121.70}^{\pm 1.31}$ & $186.60^{\pm 3.23}$ & $\textbf{0.79}^{\pm 0.00}$ & $0.97^{\pm 0.00}$ \\
			& & \CC{10}$50$ & \CC{10}$\textbf{59.0}^{\pm 0.1}$ & \CC{10}$129.95^{\pm 0.39}$ & \CC{10}$\textbf{164.38}^{\pm 2.27}$ & \CC{10}$0.74^{\pm 0.00}$ & \CC{10}$0.96^{\pm 0.01}$ \\\cmidrule{2-8}
			& \multirow{3}{*}{\rotatebox{90}{Tran.\;\;\;}} 
			& $10$ & $47.6^{\pm 0.2}$ & $214.35^{\pm 4.38}$ & $151.66^{\pm 2.62}$ & $0.66^{\pm 0.00}$ & $0.79^{\pm 0.00}$ \\
			& & $20$ & $50.0^{\pm 0.1}$ & $186.28^{\pm 3.73}$ & $140.95^{\pm 2.43}$ & $\textbf{0.71}^{\pm 0.00}$ & $0.87^{\pm 0.00}$ \\
			& & \CC{10}$50$ & \CC{10}$\textbf{56.8}^{\pm 0.2}$ & \CC{10}$\textbf{141.85}^{\pm 2.51}$ & \CC{10}$\textbf{139.82}^{\pm 1.80}$ & \CC{10}$0.67^{\pm 0.00}$ & \CC{10}$\textbf{0.88}^{\pm 0.00}$ \\
			\bottomrule
		\end{tabular}
	}
	\caption{\textbf{Ablation study} on the number of frames to predict ($P$). The model we choose is highlighted with a blue background.}
	\label{tab:abla_pad}
\end{table}
\begin{table}[!ht]
	\centering
	\resizebox{\linewidth}{!}{
		\begin{tabular}{ccc|ccccc}
			\toprule
			& & $\delta$ & Acc$\uparrow$ & $FID_{\text{tr}}\downarrow$ & $FID_{\text{te}}\downarrow$ & $Div_{\text{w}}\uparrow$  & $Div\uparrow$\\
			\midrule
			\multirow{6}{*}{\rotatebox{90}{GRAB\;\;\;\;}}& \multirow{3}{*}{\rotatebox{90}{RNN\;\;\;}} 
			& $0.005$ & $78.4^{\pm 0.4}$ & $152.74^{\pm 2.33}$ & $\textbf{21.90}^{\pm 1.12}$  & $\textbf{1.17}^{\pm 0.01}$ & $1.37^{\pm 0.01}$ \\
			& & \CC{10}$0.015$ & \CC{10}$\textbf{92.6}^{\pm 0.6}$ & \CC{10}$\textbf{44.59}^{\pm 1.39}$ & \CC{10}$38.03^{\pm 1.49}$ & \CC{10}$1.10^{\pm 0.01}$ & \CC{10}$1.37^{\pm 0.01}$ \\
			& & $0.025$ & $88.1^{\pm 0.4}$ & $47.93^{\pm 1.83}$ & $31.55^{\pm 1.26}$ & $1.07^{\pm 0.01}$ & $1.37^{\pm 0.01}$ \\\cmidrule{2-8}
			& \multirow{3}{*}{\rotatebox{90}{Tran.\;\;\;}} 
			& $0.005$ & $51.1^{\pm 0.4}$ & $276.29^{\pm 5.44}$ & $111.65^{\pm 3.22}$ & $1.02^{\pm 0.01}$ & $1.08^{\pm 0.01}$ \\
			& & \CC{10}$0.015$ & \CC{10}$\textbf{85.5}^{\pm 1.2}$ & \CC{10}$48.58^{\pm 3.05}$ & \CC{10}$\textbf{25.72}^{\pm 2.16}$ & \CC{10}$\textbf{1.05}^{\pm 0.01}$ & \CC{10}$1.08^{\pm 0.01}$ \\
			& & $0.025$ & $82.7^{\pm 1.0}$ & $\textbf{43.04}^{\pm 2.46}$ & $27.75^{\pm 1.65}$ & $1.03^{\pm 0.01}$ & $1.08^{\pm 0.01}$ \\
			\bottomrule
			
			\multirow{6}{*}{\rotatebox{90}{NTU\;\;\;\;}}& \multirow{3}{*}{\rotatebox{90}{RNN\;\;\;}} 
			& $0.005$ & $78.8^{\pm 0.2}$ & $71.22^{\pm 0.63}$ & $182.79^{\pm 1.97}$ & $\textbf{1.39}^{\pm 0.00}$ & $2.20^{\pm 0.01}$ \\
			& & $0.015$ & $\textbf{78.9}^{\pm 0.2}$ & $\textbf{51.19}^{\pm 0.54}$ & $129.20^{\pm 1.52}$ & $1.29^{\pm 0.00}$ & $2.20^{\pm 0.01}$ \\
			& & \CC{10}$0.025$ & \CC{10}$76.0^{\pm 0.2}$ & \CC{10}$72.18^{\pm 0.93}$ & \CC{10}$\textbf{111.01}^{\pm 1.28}$ & \CC{10}$1.25^{\pm 0.00}$ & \CC{10}$2.20^{\pm 0.01}$ \\\cmidrule{2-8}
			& \multirow{3}{*}{\rotatebox{90}{Tran.\;\;\;}} 
			& $0.005$ & $62.2^{\pm 0.2}$ & $256.95^{\pm 1.55}$ & $257.44^{\pm 1.90}$ & $\textbf{1.80}^{\pm 0.01}$ & $2.19^{\pm 0.01}$ \\
			& & $0.015$ & $\textbf{72.7}^{\pm 0.2}$ & $\textbf{77.45}^{\pm 0.76}$ & $158.81^{\pm 2.07}$ & $1.38^{\pm 0.00}$ & $2.19^{\pm 0.01}$ \\
			& & \CC{10}$0.025$ & \CC{10}$71.3^{\pm 0.2}$ & \CC{10}$83.14^{\pm 1.74}$ & \CC{10}$\textbf{114.62}^{\pm 0.93}$ & \CC{10}$1.25^{\pm 0.00}$ & \CC{10}$2.19^{\pm 0.01}$ \\
			\bottomrule
			
			\multirow{6}{*}{\rotatebox{90}{BABEL\;}}& \multirow{3}{*}{\rotatebox{90}{RNN\;\;\;}} 
			&  $0.005$ & $45.4^{\pm 0.2}$ & $45.00^{\pm 0.10}$ & $43.40^{\pm 0.30}$ & $1.57^{\pm 0.00}$ & $1.74^{\pm 0.00}$\\
			& & $0.015$ & $\textbf{51.1}^{\pm 0.3}$ & $26.00^{\pm 0.24}$ & $25.80^{\pm 0.34}$ & $\textbf{1.43}^{\pm 0.00}$ & $1.74^{\pm 0.00}$\\
			& & \CC{10}$0.025$ & \CC{10}$49.6^{\pm 0.4}$ & \CC{10}$\textbf{22.50}^{\pm 0.27}$ & \CC{10}$\textbf{22.40}^{\pm 0.36}$ & \CC{10}$1.35^{\pm 0.00}$ & \CC{10}$1.74^{\pm 0.00}$\\\cmidrule{2-8}
			& \multirow{3}{*}{\rotatebox{90}{Tran.\;\;\;}} 
			& $0.005$ & $31.7^{\pm 0.1}$ & $68.54^{\pm 0.32}$ & $64.92^{\pm 0.27}$ & $\textbf{1.78}^{\pm 0.01}$ & $1.82^{\pm 0.01}$\\
			& & $0.015$ & $\textbf{39.5}^{\pm 0.2}$ & $31.04^{\pm 0.20}$ & $30.18^{\pm 0.34}$ & $1.58^{\pm 0.00}$ & $1.82^{\pm 0.01}$\\
			& & \CC{10}$0.025$ & \CC{10}$\textbf{39.5}^{\pm 0.3}$ & \CC{10}$\textbf{20.02}^{\pm 0.24}$ & \CC{10}$\textbf{19.41}^{\pm 0.35}$ & \CC{10}$1.39^{\pm 0.00}$ & \CC{10}$1.82^{\pm 0.01}$ \\
			\bottomrule
			
			\multirow{6}{*}{\rotatebox{90}{HAct\;}}& \multirow{3}{*}{\rotatebox{90}{RNN\;\;\;}} 
			& \CC{10}$0.01$ & \CC{10}$\textbf{59.0}^{\pm 0.1}$ & \CC{10}$\textbf{129.95}^{\pm 0.39}$ & \CC{10}$164.38^{\pm 2.27}$ & \CC{10}$\textbf{0.74}^{\pm 0.00}$ & \CC{10}$0.96^{\pm 0.00}$ \\
			& & $0.015$ & $54.2^{\pm 0.1}$ & $215.68^{\pm 2.99}$ & $\textbf{158.94}^{\pm 2.09}$ & $0.71^{\pm 0.00}$ & $0.96^{\pm 0.01}$ \\
			& & $0.02$ & $50.4^{\pm 0.1}$ & $283.92^{\pm 4.27}$ & $179.30^{\pm 2.49}$ & $0.68^{\pm 0.00}$ & $0.96^{\pm 0.01}$ \\\cmidrule{2-8}
			& \multirow{3}{*}{\rotatebox{90}{Tran.\;\;\;}} 
			& $0.01$ & $50.9^{\pm 0.1}$ & $402.58^{\pm 7.32}$ & $759.53^{\pm 10.26}$ & $\textbf{0.72}^{\pm 0.00}$ & $0.88^{\pm 0.00}$ \\
			& & $0.015$ & $\textbf{59.7}^{\pm 0.2}$ & $\textbf{97.92}^{\pm 1.37}$ & $232.52^{\pm 3.90}$ & $0.69^{\pm 0.00}$ & $0.88^{\pm 0.00}$ \\
			& & \CC{10}$0.02$ & \CC{10}$56.8^{\pm 0.2}$ & \CC{10}$141.85^{\pm 2.51}$ & \CC{10}$\textbf{139.82}^{\pm 1.80}$ & \CC{10}$0.67^{\pm 0.00}$ & \CC{10}$0.88^{\pm 0.00}$ \\
			\bottomrule
		\end{tabular}
	}
	\caption{\textbf{Ablation study} on the stop threshold ($\delta$). The model we choose is highlighted with a blue background.}
	\label{tab:abla_th}
\end{table}

\section{Additional Details}
\textbf{Smoothness prior.} We use $L=10$ frames for our smoothness prior, which means that we take the last 10 poses of the history and the first 10 ones of the future motion to form a sequence of length 20. We set the number of DCT bases ($M$) to 5. 

\textbf{Variable-length motion prediction.} To enable predicting variable-length sequences, during training, we make the model generate $P$ additional frames, and supervise these frames with the last pose of the future to encourage the model to generate static poses (i.e., the last pose of the ground-truth motion) after reaching the motion end. We validate the number of additional frames to generate~($P$). The numerical results are shown in Table~\ref{tab:abla_pad}.

During test time, we stop the prediction when the variance of the last $Q$ consecutive frames falls below a threshold ($\delta$). For all our experiments, $Q$ is set to 5, and the  stopping threshold $\delta$ ranges from $0.005$ to $0.025$. In Table~\ref{tab:abla_th}, we validate this threshold for the different datasets.

\begin{table}[!ht]
	\centering
	\resizebox{\linewidth}{!}{
		\begin{tabular}{ccc|ccccc}
			\toprule
			&\multicolumn{2}{c|}{Method} & Acc$\uparrow$ & $FID_{\text{tr}}\downarrow$ & $FID_{\text{te}}\downarrow$ & $Div_{\text{w}}\uparrow$  & $Div\uparrow$\\
			\midrule
			\multirow{4}{*}{\rotatebox{90}{GRAB\;\;\;\;}}& \multirow{2}{*}{\rotatebox{90}{RNN\;}} 
			& stop sign & $66.5^{\pm 1.0}$ & $211.11^{\pm 6.89}$ & $\textbf{31.49}^{\pm 3.35}$ & $\textbf{1.31}^{\pm 0.01}$ & $\textbf{1.44}^{\pm 0.01}$\\\cmidrule{3-8}
			& & \CC{10}padding & \CC{10}$\textbf{92.6}^{\pm 0.6}$ & \CC{10}$\textbf{44.59}^{\pm 1.39}$ & \CC{10}$38.03^{\pm 1.49}$ & \CC{10}$1.10^{\pm 0.01}$ & \CC{10}$1.37^{\pm 0.01}$ \\\cmidrule{2-8}
			& \multirow{2}{*}{\rotatebox{90}{Tran.\;}} 
			& stop sign & $76.3^{\pm 1.4}$ & $101.07^{\pm 4.93}$ & $\textbf{16.02}^{\pm 1.03}$ & $0.05^{\pm 0.00}$ & $0.27^{\pm 0.00}$ \\\cmidrule{3-8}
			& & \CC{10}padding & \CC{10}$\textbf{85.5}^{\pm 1.2}$ & \CC{10}$\textbf{48.58}^{3.05}$ & \CC{10}$25.72^{\pm 2.16}$ & \CC{10}$\textbf{1.05}^{\pm 0.01}$ & \CC{10}$\textbf{1.08}^{\pm 0.01}$ \\
			\bottomrule
			
			\multirow{4}{*}{\rotatebox{90}{NTU\;\;\;}}& \multirow{2}{*}{\rotatebox{90}{RNN\;\;}} 
			& stop sign & $29.2^{\pm 0.1}$ & $1188.50^{\pm 2.72}$ & $1234.79^{\pm 3.99}$ & $1.13^{\pm 0.00}$ & $1.50^{\pm 0.00}$ \\\cmidrule{3-8}
			& & \CC{10}padding & \CC{10}$\textbf{76.0}^{\pm 0.2}$ & \CC{10}$\textbf{72.18}^{\pm 0.93}$  & \CC{10}$\textbf{111.01}^{\pm 1.28}$ & \CC{10}$\textbf{1.25}^{\pm 0.00}$ & \CC{10}$\textbf{2.20}^{\pm 0.00}$ \\\cmidrule{2-8}
			& \multirow{2}{*}{\rotatebox{90}{Tran.\;}} 
			& stop sign & $\textbf{77.7}^{\pm 0.2}$ & $259.22^{\pm 2.04}$ & $358.79^{\pm 2.68}$ & $0.00^{\pm 0.00}$ & $0.01^{\pm 0.00}$ \\
			\cmidrule{3-8}
			& & \CC{10}padding & \CC{10}$71.3^{\pm 0.2}$ & \CC{10}$\textbf{83.14}^{\pm 1.74}$  & \CC{10}$\textbf{114.62}^{\pm 0.93}$ & \CC{10}$\textbf{1.25}^{\pm 0.00}$ & \CC{10}$\textbf{2.19}^{\pm 0.01}$ \\
			\bottomrule
			
			\multirow{4}{*}{\rotatebox{90}{BABEL\;}}& \multirow{2}{*}{\rotatebox{90}{RNN\;}} 
			& stop sign & $13.3^{\pm 0.1}$ & $162.76^{\pm 3.23}$ & $164.01^{\pm 3.25}$ & $0.93^{\pm 0.00}$ & $\textbf{2.14}^{\pm 0.01}$ \\
			\cmidrule{3-8}
			& & \CC{10}padding & \CC{10}$\textbf{49.6}^{\pm 0.4}$ & \CC{10}$\textbf{22.50}^{\pm 0.27}$ & \CC{10}$\textbf{22.40}^{\pm 0.36}$ & \CC{10}$\textbf{1.35}^{\pm 0.00}$ & \CC{10}$1.74^{\pm 0.00}$\\\cmidrule{2-8}
			& \multirow{2}{*}{\rotatebox{90}{Tran.\;}} 
			& stop sign & $15.2^{\pm 0.3}$ & $25.43^{\pm 0.29}$ & $22.26^{\pm 0.44}$ & $0.46^{\pm 0.01}$ & $1.23^{\pm 0.00}$ \\
			\cmidrule{3-8}
			& & \CC{10}padding & \CC{10}$\textbf{39.5}^{\pm 0.3}$ & \CC{10}$\textbf{20.02}^{\pm 0.24}$ & \CC{10}$\textbf{19.41}^{\pm 0.35}$ & \CC{10}$\textbf{1.39}^{\pm 0.00}$ & \CC{10}$\textbf{1.82}^{\pm 0.01}$\\
			\bottomrule
			
			\multirow{4}{*}{\rotatebox{90}{HAct\;}}& \multirow{2}{*}{\rotatebox{90}{RNN\;}} 
			& stop sign & $43.7^{\pm 0.2}$ & $\textbf{103.61}^{\pm 1.94}$ & $222.14^{\pm 6.66}$ & $\textbf{0.76}^{\pm 0.00}$ & $0.84^{\pm 0.00}$ \\
			\cmidrule{3-8}
			& & \CC{10}padding & \CC{10}$\textbf{59.0}^{\pm 0.1}$ & \CC{10}$129.95^{\pm 0.39}$ & \CC{10}$\textbf{164.38}^{\pm 2.27}$ & \CC{10}$0.74^{\pm 0.00}$ & \CC{10}$\textbf{0.96}^{\pm 0.00}$ \\\cmidrule{2-8}
			& \multirow{2}{*}{\rotatebox{90}{Tran.\;}} 
			& stop sign & $47.7^{\pm 0.2}$ & $\textbf{112.33}^{\pm 3.10}$ & $221.10^{\pm 5.62}$ & $0.04^{\pm 0.00}$ & $0.05^{\pm 0.00}$ \\
			\cmidrule{3-8}
			& & \CC{10}padding & \CC{10}$\textbf{56.8}^{\pm 0.2}$ & \CC{10}$141.85^{\pm 2.51}$ & \CC{10}$\textbf{139.82}^{\pm 1.80}$ & \CC{10}$\textbf{0.67}^{\pm 0.00}$ & \CC{10}$\textbf{0.88}^{\pm 0.00}$ \\
			\bottomrule
		\end{tabular}
	}
	\caption{\textbf{Ablation study} on generating variable-length future motions. The model we choose is highlighted with a blue background.}
	\label{tab:stop_sign}
\end{table}
As an alternative to the above-mentioned padding approach to variable-length prediction, we draw inspiration from the NLP literature. However, while the standard strategy for variable-length outputs in NLP is to predict a stop token, there is no such ``stop token'' for human pose. Thus, we make the model output one more value together with the human pose at each prediction step. The additional value is supervised by a binary label indicating whether the motion ends or not. Such a value is then used as a ``stop sign'' during testing. We compare the results of two the strategies described above (stop sign vs padding) in Table~\ref{tab:stop_sign}. In general, the models with padding perform better than those with the stop sign.

\section{Stopping Strategy for Action2Motion}
Due to the jitter produced by the Action2Motion~\cite{guo2020action2motion} model, we have observed our variance-based stopping criterion to be sub-optimal. We therefore test a stopping strategy based on the difference between consecutive frames. That is, if the difference between the latest 2 frames falls below certain threshold $\delta$, the model stops predicting the future motions. In Table~\ref{tab:abla_stop}, we compare the performance of the variance-based strategy and the adjacent frames' difference-based one with different stopping thresholds. Overall, the stopping strategy based on the difference of adjacent frames outperforms the variance-based one. In the main paper, we report the values highlighted with a blue background.

\section{Details of the Datasets}
We evaluate our method on four different datasets, i.e., GRAB~\cite{GRAB:2020,Brahmbhatt_2019_CVPR}, NTU RGB-D~\cite{shahroudy2016ntu,liu2019ntu}, BABEL~\cite{BABEL:CVPR:2021} and HumanAct12~\cite{guo2020action2motion}.

\textbf{License.} These four datasets all are for non-commercial scientific research use only. For the details of their individual licenses, please refer to their official websites:
\begin{itemize}
	\item GRAB: \url{https://grab.is.tue.mpg.de/}
	\item NTU RGB-D: \url{http://rose1.ntu.edu.sg/Datasets/actionRecognition.asp}
	\item BABEL:
	\url{https://babel.is.tue.mpg.de/data.html}
	\item HumanAct12: \url{https://jimmyzou.github.io/publication/2020-PHSPDataset}
\end{itemize}

\section{Additional Qualitative Results}
We provide additional qualitative results in the supplementary video.

\begin{table}[!ht]
	\centering
	\resizebox{\linewidth}{!}{
		\begin{tabular}{ccc|ccccc}
			\toprule
			& Stop criterion & $\delta$ & Acc$\uparrow$ & $FID_{\text{tr}}\downarrow$ & $FID_{\text{te}}\downarrow$ & $Div_{\text{w}}\uparrow$  & $Div\uparrow$\\
			\midrule
			\multirow{6}{*}{\rotatebox{90}{GRAB\;\;\;\;}}& \multirow{3}{*}{adjacent frames} 
			& $0.01$ &$64.3^{\pm 1.2}$ & $163.85^{\pm 10.14}$ & $56.38^{\pm 7.02}$ & $0.40^{\pm 0.00}$ & $0.76^{\pm 0.01}$\\
			& & $0.02$ &$64.7^{\pm 1.2}$ & $159.14^{\pm 9.82}$ & $54.35^{\pm 6.93}$ & $0.43^{\pm 0.00}$ & $0.76^{\pm 0.01}$\\
			& & \CC{10}$0.04$ &\CC{10}$\textbf{70.6}^{\pm 1.3}$ & \CC{10}$\textbf{80.22}^{\pm 6.64}$ & \CC{10}$\textbf{47.81}^{\pm 1.09}$ & \CC{10}$\textbf{0.50}^{\pm 0.00}$ & \CC{10}$0.76^{\pm 0.01}$\\\cmidrule{2-8}
			& \multirow{3}{*}{variance based} & $0.01$ &$64.3^{\pm 1.2}$ & $163.85^{\pm 10.14}$ & $56.38^{\pm 7.02}$ & $0.40^{\pm 0.00}$ & $0.76^{\pm 0.01}$\\
			& & $0.02$ &$64.3^{\pm 1.2}$ & $163.69^{\pm 10.12}$ & $56.34^{\pm 7.02}$ & $0.40^{\pm 0.00}$ & $0.76^{\pm 0.01}$\\
			& & $0.04$ &$69.8^{\pm 1.2}$ & $100.76^{\pm 7.78}$ & $35.20^{\pm 3.04}$ & $0.49^{\pm 0.01}$ & $0.76^{\pm 0.01}$\\
			\bottomrule
			
			\multirow{6}{*}{\rotatebox{90}{NTU\;\;\;\;}}& \multirow{3}{*}{adjacent frames}
			& $0.01$ &$60.5^{\pm 0.2}$ & $354.92^{\pm 1.77}$ & $442.99^{\pm 3.80}$ & $0.67^{\pm 0.01}$ & $1.19^{\pm 0.01}$\\
			& & $0.02$ &$61.9^{\pm 0.2}$ & $316.37^{\pm 1.87}$ & $415.69^{\pm 3.14}$ & $0.68^{\pm 0.01}$ & $1.19^{\pm 0.01}$\\
			& & \CC{10}$0.04$ &\CC{10}$\textbf{66.3}^{\pm 0.2}$ & \CC{10}$\textbf{144.98}^{\pm 2.44}$ & \CC{10}$\textbf{113.61}^{\pm 0.84}$ & \CC{10}$\textbf{0.75}^{\pm 0.01}$ & \CC{10}$1.19^{\pm 0.01}$\\\cmidrule{2-8}
			& \multirow{3}{*}{variance based} & $0.01$ &$60.5^{\pm 0.2}$ & $354.86^{\pm 1.75}$ & $442.80^{\pm 3.78}$ & $0.67^{\pm 0.01}$ & $1.19^{\pm 0.01}$\\
			& & $0.02$ &$60.6^{\pm 0.2}$ & $350.17^{\pm 1.80}$ & $439.70^{\pm 3.67}$ & $0.67^{\pm 0.01}$ & $1.19^{\pm 0.01}$\\
			& & $0.04$ &$64.3^{\pm 0.2}$ & $177.21^{\pm 0.96}$ & $236.82^{\pm 2.28}$ & $0.70^{\pm 0.01}$ & $1.19^{\pm 0.01}$\\
			\bottomrule
			
			\multirow{6}{*}{\rotatebox{90}{BABEL\;\;}}& \multirow{3}{*}{adjacent frames}
			& $0.01$ & $\textbf{15.9}^{\pm 0.2}$ & $62.95^{\pm 0.40}$ & $57.39^{\pm 0.31}$ & $0.76^{\pm 0.01}$ & $1.10^{\pm 0.01}$\\
			& & $0.02$ & $\textbf{15.9}^{\pm 0.2}$ & $62.81^{\pm 0.39}$ & $57.26^{\pm 0.30}$ & $0.76^{\pm 0.01}$ & $1.10^{\pm 0.01}$ \\
			& & \CC{10}$0.04$ & \CC{10}$14.8^{\pm 0.2}$ & \CC{10}$\textbf{42.02}^{\pm 0.40}$ & \CC{10}$\textbf{37.41}^{\pm 0.47}$ & \CC{10}$\textbf{0.79}^{\pm 0.01}$ & \CC{10}$1.10^{\pm 0.01}$ \\\cmidrule{2-8}
			& \multirow{3}{*}{variance based} 
			& $0.01$ & $\textbf{15.9}^{\pm 0.2}$ & $62.95^{\pm 0.40}$ & $57.39^{\pm 0.31}$ & $0.76^{\pm 0.01}$ & $1.10^{\pm 0.01}$ \\
			& & $0.02$ & $\textbf{15.9}^{\pm 0.2}$ & $62.93^{\pm 0.40}$ & $57.38^{\pm 0.31}$ & $0.76^{\pm 0.01}$ & $1.10^{\pm 0.01}$ \\
			& & $0.04$ & $15.5^{\pm 0.2}$ & $56.48^{\pm 0.20}$ & $51.16^{\pm 0.24}$ & $0.76^{\pm 0.01}$ & $1.10^{\pm 0.01}$ \\
			\bottomrule
			
			\multirow{6}{*}{\rotatebox{90}{HAct\;\;}}& \multirow{3}{*}{adjacent frames}
			& $0.01$ &$25.1^{\pm 0.2}$ & $388.98^{\pm 15.31}$ & $585.59^{\pm 21.07}$ & $0.26^{\pm 0.00}$ & $0.60^{\pm 0.01}$\\
			& & \CC{10}$0.02$ &\CC{10}$24.5^{\pm 0.1}$ & \CC{10}$245.34^{\pm 7.13}$ & \CC{10}$298.06^{\pm 10.79}$ & \CC{10}$\textbf{0.31}^{\pm 0.00}$ & \CC{10}$0.60^{\pm 0.01}$\\
			& & $0.04$ &$20.5^{\pm 0.2}$ & $292.83^{\pm 6.37}$ & $\textbf{134.47}^{\pm 2.69}$ & $\textbf{0.31}^{\pm 0.00}$ & $0.60^{\pm 0.01}$\\\cmidrule{2-8}
			& \multirow{3}{*}{variance based} & $0.01$ &$25.0^{\pm 0.2}$ & $395.59^{\pm 15.56}$ & $598.76^{\pm 21.49}$ & $0.25^{\pm 0.00}$ & $0.60^{\pm 0.01}$\\
			& & $0.02$ &$\textbf{25.2}^{\pm 0.1}$ & $337.77^{\pm 12.19}$ & $486.46^{\pm 16.87}$ & $0.28^{\pm 0.00}$ & $0.60^{\pm 0.01}$\\
			& & $0.04$ &$22.6^{\pm 0.2}$ & $\textbf{224.47}^{\pm 5.67}$ & $150.82^{\pm 4.00}$ & $\textbf{0.31}^{\pm 0.00}$ & $0.60^{\pm 0.01}$\\
			\bottomrule
		\end{tabular}
	}
	\caption{\textbf{Ablation study} on the stopping criterion for Action2Motion~\cite{guo2020action2motion}. The model we report results with is highlighted with a blue background.}
	\label{tab:abla_stop}
\end{table}

\end{document}